\pdfoutput=1

\documentclass[11pt]{article}

\usepackage[]{EMNLP2022}

\usepackage{times}
\usepackage{latexsym}

\usepackage[T1]{fontenc}

\usepackage[utf8]{inputenc}

\usepackage{microtype}

\usepackage{inconsolata}

\usepackage{booktabs}
\usepackage{times}
\usepackage{latexsym}
\usepackage{amsmath}
\usepackage{float}
\usepackage{footnote}
\usepackage{enumitem}
\usepackage{bm}
\usepackage{arydshln}
\usepackage{booktabs}
\usepackage{multicol}
\usepackage{multirow}
\usepackage{amssymb}
\usepackage{color}
\usepackage{xcolor}     
\usepackage{colortbl}
\usepackage{bbding}
\usepackage{makecell}
\usepackage{mathtools}
\usepackage{subfigure}
\usepackage{CJKutf8}
\usepackage{makecell}
\usepackage[T1]{fontenc}

\usepackage[utf8]{inputenc}
\usepackage{fontawesome}
\usepackage{fdsymbol}

\usepackage{bbding}
\usepackage[most]{tcolorbox}
\definecolor{my_darkblue}{rgb}{0.36, 0.54, 0.66}
\definecolor{bar_purple}{rgb}{0.87, 0.45, 1.0}
\definecolor{bar_green}{rgb}{0.01, 0.75, 0.24}
\definecolor{icon_purple}{rgb}{0.75, 0.0, 1.0}
\definecolor{icon_green}{rgb}{0.0, 0.5, 0.0}
\definecolor{myblue}{rgb}{0.9, 0.1, 0.94}
\definecolor{mygreen}{rgb}{0.64, 0.56, 0.88}
\definecolor{myyellow}{rgb}{0.98, 0.94, 0.75}
\definecolor{mygreen}{rgb}{0.68, 0.9, 0.6}
\definecolor{myorange}{rgb}{1.0, 0.49, 0.0}	
\definecolor{bittersweet}{rgb}{1.0, 0.44, 0.37}
\definecolor{gray}{rgb}{0.75, 0.75, 0.75}

\usepackage{microtype}
\usepackage{arydshln}
\usepackage{multirow}

\newenvironment{itemize*}%
 {\leftmargini=10pt\begin{itemize}%
  \setlength{\itemsep}{0pt}%
  \setlength{\parskip}{0pt}%
  }%
 {\end{itemize}}
\newenvironment{enumerate*}%
 {\begin{enumerate}%
  \setlength{\itemsep}{0pt}%
  \setlength{\parskip}{0pt}}%
 {\end{enumerate}}

\title{\textit{Polyglot Prompting}: Multilingual Multitask
Prompt Training}

\author{Jinlan Fu \\
  NUS \\
  \text{jinlan@nus.edu.sg} \\\And
  See-Kiong Ng \\
  NUS \\
  \text{seekiong@nus.edu.sg} \\\And
  Pengfei Liu \\
  CMU \& Inspired Cognition\\
  \text{stefanpengfei@gmail.com} \\}

\begin{document}
\maketitle
\begin{abstract}
This paper aims for a potential architectural improvement for multilingual learning and asks: \textit{Can different tasks from different languages be modeled in a monolithic framework, i.e. without any task/language-specific module}? 
The benefit of achieving this could open new doors for future multilingual research, including allowing systems trained on low resources to be further assisted by other languages as well as other tasks.
We approach this goal by developing a learning framework named \textit{Polyglot Prompting} to exploit prompting methods for learning a unified semantic space for different languages and tasks with multilingual prompt engineering.
We performed a comprehensive evaluation of $6$ tasks, namely topic classification, sentiment classification, named entity recognition, question answering, natural language inference, and summarization, covering $24$ datasets and $49$ languages.  The experimental results demonstrated the efficacy of multilingual multitask prompt-based learning and led to inspiring observations. 
We also present an interpretable multilingual evaluation methodology and show how the proposed framework, multilingual multitask prompt training, works.
We release all datasets prompted in the best setting and code. 
\footnote{\url{https://github.com/jinlanfu/Polyglot_Prompt}}

\end{abstract}

\begin{CJK*}{UTF8}{gbsn}

\section{Introduction}
\label{sec:intro}

\begin{figure}[!th]
    \centering
    \includegraphics[width=0.95\linewidth]{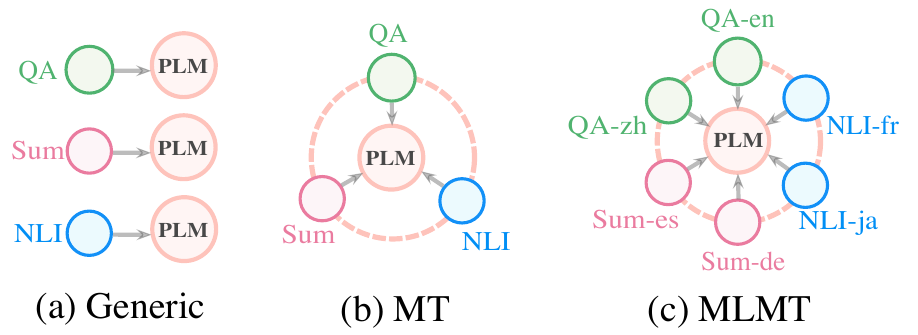}
    \vspace{-7pt}
    \caption{
Application of prompt technology in three different scenarios: Generic, Multitask (MT), Multilingual Multitask (MLMT).  \textit{QA}, \textit{Sum}, and \textit{NLI} represent different tasks, namely question answering, summarization, and natural language inference here. PLM represents pre-trained language model, and ``zh'',   ``en'', ``fr'', ``ja'', ``de'',``es''  denote different languages.
}
	\label{fig:mtl-overview}
\end{figure}

The emergence of multilingual pre-trained language models~\cite{linting2021mt5,yinhan2020mbart,alexis2020xlmr,conneau2019xlm} enables different languages to be represented in a unified semantic space.  As a result, a fine-tuned model of a data-rich language such as English can achieve decent transfer (e.g., zero-shot) performance in geographically, syntactically, or phonetically similar languages~\cite{chaitanya2017learning}. The insufficient features learned by  languages under lower-resource settings can thus be compensated through the higher-resource languages shared with them.

Despite the preliminary success in the low-resource scenarios using  shared knowledge across languages in multilingual language models~\cite{wang2019clbert,liu2019icl,k2020mbertstudy}, the cross-lingual transfers have mostly occurred only within the boundary of the same task or similar tasks.
\textit{Can more conceptually-diverse tasks from different 
languages communicate together?}
While researchers have made some preliminary progress towards this direction \cite{yang2016multi,lin-etal-2018-multilingual,tahmid2021xlsum,dong-etal-2015-multi,mahabadi2021parameter},  the scope of the ``different'' tasks had remained relatively narrow, such as limiting to tasks with the same sequence labeling form \cite{yang2016multi}  (e.g., named entity recognition, chunking, and part-of-speech), or different domains for the same task \cite{wang-etal-2020-multi}.

Unifying different tasks into one framework can be challenging if we are to avoid introducing additional task-specific parameterized modules. Recently, the success of the prompting methods~\cite{liu2021prompt,victor2021t0} has provided us with new clues on unifying different tasks in the same framework without task-specific parameters by formulating all tasks as a pre-training problem with various frameworks such as the mask language model~\cite{devlin2019bert,alexis2020xlmr} or the encoder-decoder model~\cite{linting2021mt5,colin2020t5,yinhan2020mbart,zewen2021mt6}.

In this paper, we leverage prompt techniques to cross the boundaries of different tasks and languages so that multiple tasks in different languages can be placed in a \textit{monolithic} framework (as shown in Fig.~\ref{fig:mtl-overview}-(c) as opposed to single task single language (Fig.~\ref{fig:mtl-overview}-(a)) as well as multiple task single language learning (Fig.~\ref{fig:mtl-overview}-(b))) to benefit from one another without requiring any task/language-specific modules.

We name this multilingual multitask training model as \textit{Polyglot Prompting (PolyPrompt)}. Different tasks from different languages can then be seamlessly connected together by being reformulated as pre-training tasks.
Architecturally, we choose the encoder-decoder pre-training framework so that more NLP tasks could be unified, as compared to other architectures such as the mask language model that favors classification-based tasks.
Our explorations in this paper are driven by following research questions:

\noindent
\textbf{Q1:} \textit{Can different tasks from different languages benefit from each other by a monolithic framework?}
If the answer is ``yes'', can the performance be further improved by introducing more high-resource datasets that are more readily available?\footnote{It is relatively easy for us to obtain the training set of relevant tasks in real scenarios. Therefore, the purpose of this research question is to explore whether the data that is relatively easy to obtain from relevant tasks can bring benefits after being introduced into the multitask learning framework.}
We develop \textit{PolyPrompt}, a new multitask multilingual learning framework, and study the performance influenced by the introduction of 17 high-resource datasets. (Sec.~\ref{sec:exp1})

\noindent
\textbf{Q2:} \textit{
Can \textit{PolyPrompt} benefit all languages in different datasets? If not, how do different characteristics of datasets and languages affect the performance of \textit{PolyPrompt}?}
We try to give answers by designing a multilingual interpretable evaluation methodology \cite{fu2021interpteval,liu2021explainaboard} to analyze the strengths and weaknesses of the unified framework for different tasks, datasets, and languages. (Sec.~\ref{sec:interpret})

\noindent
\textbf{Q3:} \textit{ What makes a good prompt for multilingual multitask prompt training?}
Applying the prompting method to a multilingual multitasking setting requires considering various difficulties of prompt engineering in the linguistic dimension.
We study two aspects of the prompt designs for\textit{PolyPrompt}: the language choice of prompt templates and the uniformity of prompt templates across tasks. (Sec.~\ref{sec:exp3})

The main observations are listed in Sec.~\ref{sec:conclusion}. Below, we summarize the main contributions.
(1) To the best of our knowledge,
    this is the first architectural exploration for the learning of multiple conceptually-different tasks (e.g., classification, question answering and text generation) and multiple diverse languages, which relies solely on a monolithic model.
(2) We introduce the concept of \textit{multilingual prompt engineering} and provide empirical insights on what makes a good multilingual prompt.
(3) We have conducted extensive experiments for \textit{in-language training}, \textit{cross-lingual zero-shot transfer}, and \textit{cross-task \& cross-lingual zero-shot transfer} scenarios, and designed an interpretable multilingual evaluation methodology to understand how multitask multilingual prompting works, which leads to interesting observations (Sec.~\ref{sec:interpret}).

\begin{figure*}[!ht]
    \centering
    \includegraphics[width=0.95\linewidth]{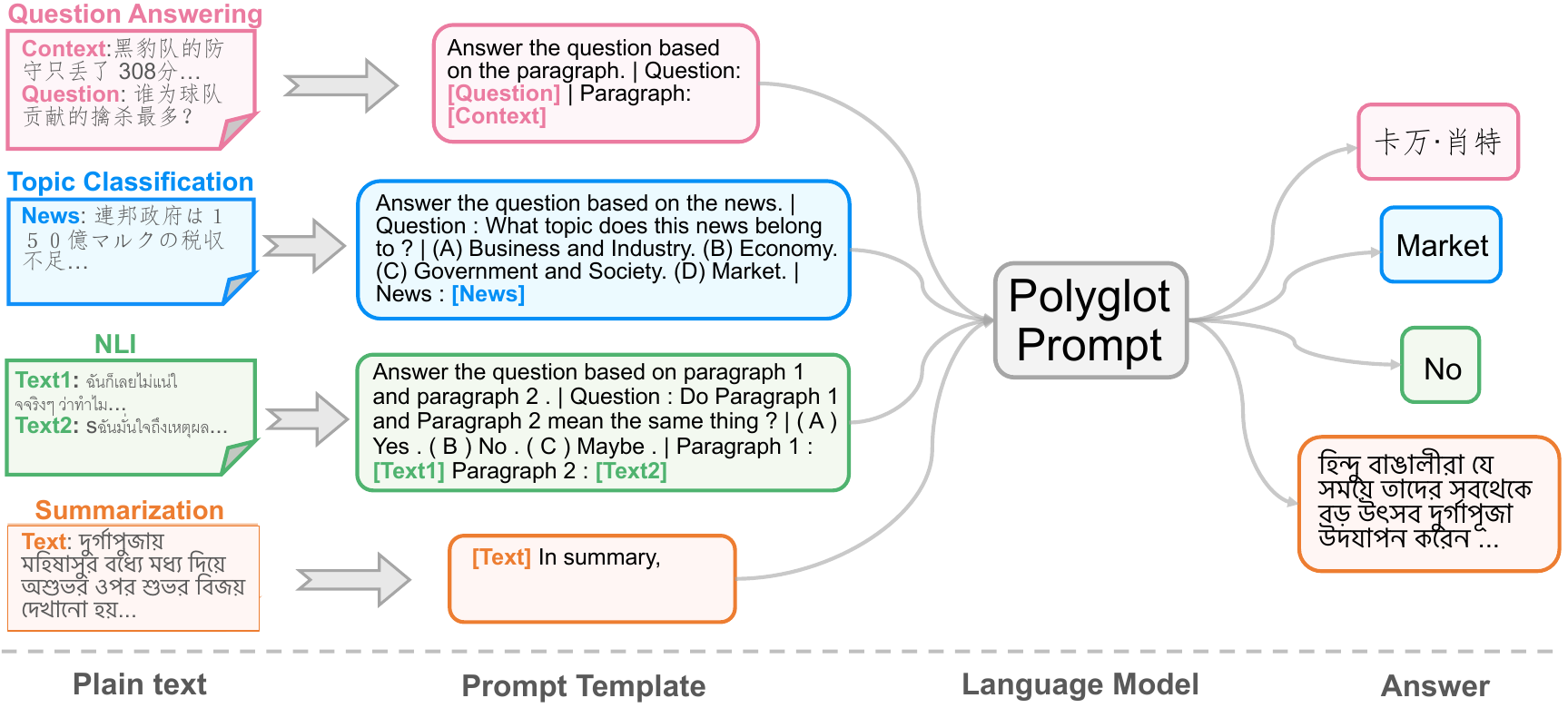}
    \vspace{-7pt}
    \caption{
    The proposed \textit{PolyPrompt} framework for  multilingual multitask prompt training. 
    }
    \label{fig:model_framework}
\end{figure*}

\section{Related Work}

\paragraph{Multitask \& Multilingual Learning}
The  developments of neural networks have made it easier to share information across tasks or languages. As such, in the past few years, there has been much work on multitask learning within the same language \cite{liu-etal-2015-representation,liu2016recurrent,sogaard-goldberg-2016-deep,kumar2016ask}, or multilingual learning in the same or similar types of tasks such as sequence labeling \cite{yang2016multi,lin-etal-2018-multilingual} and machine translation \cite{dong-etal-2015-multi,mahabadi2021parameter,hu2021deep}.  
However, the task of learning different languages and tasks simultaneously in a unified learning framework without task- or language-dependent parameters has remained unexplored.

\paragraph{Prompting Methods}

Prompting is a technique that aims to make better use of pre-trained knowledge by reformulating tasks at hand accordingly \cite{liu2021prompt}  and thus allowing us to do more with one model by unifying signals cross tasks \cite{victor2021t0}, languages \cite{multilingual2021prompt}, even modality \cite{jinming2021multimodal}.
In this paper, we expand what a system can do by proposing multilingual multitask learning with prompting methods for connecting geographically diverse languages and linguistically different tasks, thereby allowing them to leverage one another effectively.

\section{Multitask Multilingual Prompt Training}

We unify different tasks from different languages by reformulating each NLP task as a sequence-to-sequence problem~\cite{sutskeverVL14seq2seq,bahdanauCB14mt} so that they can be connected by a multilingual pre-trained language model (e.g., mT5 \cite{linting2021mt5}) that also adopts a sequence-to-sequence training objective. 
Fig.~\ref{fig:model_framework} shows the overview of our proposed framework.
Each sample from different tasks will be re-formatted as a (\textit{prompt}, \textit{answer}) pair using pre-defined \textit{templates} and then be fed into a multilingual pre-trained language model.

Formally, given a task set $\mathcal{T} = \{T_1, T_2, \cdots, T_n\}$ with $n$ tasks and corresponding prompt templates $\mathcal{K} = \{K_1, K_2, \cdots, K_n\}$ (without loss of generality, we shall assume that each task has one prompt for easy understanding).
First, we transform the samples in each task into a form that is understandable by the language model based on the predefined templates $\mathcal{K}$. 
Assume that $(\text{x}_{i,j}, \text{y}_{i,j}) \in \mathcal{Z}$ is the input and output pair for the $j$-th sample of the $i$-th task, where $\mathcal{Z}$ contains all input and output sample pairs.
The input-output pair $(\text{x}_{i,j}, \text{y}_{i,j})$ for the $j$-th sample of the $i$-th task can be converted to $(\hat{\text{x}}_{i,j}, \hat{\text{y}}_{i,j})$ through the predefined template $K_i$, which can be formulated as:
\begin{align}
(\hat{\text{x}}_{i,j}, \hat{\text{y}}_{i,j}) = K_i(\text{x}_{i,j}, \text{y}_{i,j})
\end{align}
We choose a sequence-to-sequence language model to achieve multilingual multitask prompt training, where samples from n tasks will be the input of the chosen language model.
The loss function is to maximize the log-likelihood of the output text and can be defined as:
\begin{equation}
\text{L} = \sum_{(\hat{\text{x}},\hat{\text{y}})\in (\text{X},\text{Y})} 
log(\prod_{m=1}^{|\hat{\text{y}}|}
P(\hat{\text{y}}_m|
\hat{\text{y}}_{<m}, 
\hat{\text{x}}; 
\theta) ),
\end{equation}
where $(\hat{\text{x}},\hat{\text{y}}) \in \mathcal{Z}$ represents a sequence-to-sequence text pair for any task.
$|\hat{\text{y}}|$ is the number of tokens in the decoded text, and $\hat{y}_{<m}$ is the target tokens before the time step $m$.

\section{Experiment Setup}

\subsection{Tasks \& Datasets}
\label{sec:tasks}
The datasets, tasks, and evaluation metrics studied in this work are shown in Tab.~\ref{tab:task}.
We call those datasets that provide training and test sets for multilingual multitask prompt training as \textbf{target datasets}. 
To explore the influence of introducing more high-resource English and multilingual datasets to \textit{PolyPrompt}, we present the \textbf{expanding datasets}, which only provides training datasets for multilingual multitask prompt training (we do not evaluate \textit{PolyPrompt} and its variants on expanding datasets).
Overall, we study 7 multilingual target datasets covering 4 NLP tasks (\textit{question answering, sentiment classification, topic classification, and sentence pair classification}), and 15 monolingual (English) and 2 multilingual expanding datasets covering 6 NLP tasks (\textit{text summarization, named entity recognition, and 4 tasks covered by target datasets}). Further details of the target and expanding datasets can be found in App.~\ref{sec:dataset-app}. The languages considered in this work can be seen in App.~\ref{sec:language}

\begin{table}[htb]
    \centering \scriptsize
  \renewcommand\tabcolsep{2.3pt}
    \begin{tabular}{lllcc}
    \toprule
    \textbf{Task} & \textbf{Dataset} & \textbf{Domain} & \textbf{Lang.} & \textbf{Metric} \\
    \midrule
    \multicolumn{5}{l}{\textbf{Target Datasets}} \\
    \midrule
    \multirow{3}[2]{*}{QA} & XQuAD & Wikipedia & 11    & F1 / EM \\
          & TyDiQA & Wikipedia & 9     & F1 / EM \\
          & MLQA  & Wikipedia & 7     & F1 / EM \\
    \midrule
    SC & MARC  & Amazon & 6     & Acc. \\
    \midrule
    TC & MLDOC & Reuters   & 8     & Acc. \\
    \midrule
   \multirow{2}[2]{*}{SPC} & PAWS-X & Wikipedia/Quora & 7     & Acc. \\
          & XNLI  & Misc. & 15    & Acc. \\
    \midrule
    \multicolumn{5}{l}{\textbf{Expanding Datasets }} \\
    \midrule
    Summ.  & XL-Sum & BBC  & 45    & - \\
    \midrule
    NER & Wikiann & Wikipedia & 40    & - \\
    \midrule
    \multicolumn{1}{l}{\multirow{6}[2]{*}{QA}} & SQuAD 2.0 & Wikipedia  & 1     & - \\
          & Quoref & Wikipedia  & 1     & - \\
          & NewsQA & CNN  & 1     & - \\
          & ROPES & Textbooks/Wikipedia  & 1     & - \\
          & MCTest & Misc.  & 1     & - \\
          & Social IQa & Misc.  & 1     & - \\
    \midrule
    \multicolumn{1}{l}{\multirow{3}[2]{*}{TC}} & DBpedia-2014 & DBpedia  & 1     & - \\
          & AG\_News & Reuters  & 1     & - \\
          & YATC  & Yahoo!  & 1     & - \\
    \midrule
    \multicolumn{1}{l}{\multirow{3}[2]{*}{SC}} & IMDB  & IMDb  & 1     & - \\
          & SST2  & Rotten Tomatoes  & 1     & - \\
          & ARP & Amazon  & 1     & - \\
    \midrule
    \multicolumn{1}{l}{\multirow{3}[2]{*}{SPC}} & Quora & Quora  & 1     & - \\
          & RTE   & News/Wikipedia  & 1     & - \\
          & SNLI  & Misc.  & 1     & - \\
    \bottomrule
    \end{tabular}
    \vspace{-7pt}
      \caption{The tasks and datasets studied in this work.  \textit{Lang.} and \textit{Acc.} denote ``Language'' and ``Accuracy''.
      \texttt{Summ.}, \texttt{NER}, \texttt{QA}, \texttt{TC}, \texttt{SC}, and \texttt{SPC} are 
      abbreviations for summarization, named entity recognition, question answering, topic classification, sentiment classification, and sentence pair classification.
      ``-'' denotes the task is not used for evaluation. 
      ``Misc.'' indicates that the dataset was artificially constructed or of unclear origin.
      }
  \label{tab:task}
\end{table}

\subsection{Experimental Settings}

\paragraph{Model} We list $5$ models explored in this work.

\noindent
(1) \textbf{Vanilla mT5}: In the \textit{cross-lingual zero-shot transfer} setting, mT5 is trained on the training set in English of the specific task (e.g. XNLI), while in the \textit{in-lingual training} setting, mT5 is trained on the training samples in all languages for the particular task (e.g. XNLI).

\noindent
(2) \textbf{Polyglot Prompt (PolyPrompt)}
is a standard multilingual multitask prompt training model， which is trained on 7 target datasets covering 4 NLP tasks (e.g., QA).

\noindent
(3) \textbf{PolyPrompt+Expand} is the \textit{PolyPrompt} model trained on the 7 target datasets and 15 high-resource (English) expanding datasets.

\noindent
(4) \textbf{PolyPrompt+Expand+PANX} is the \textit{PolyPrompt} trained on the 7 target datasets, 15 high-resource datasets, and a multilingual NER dataset (PANX). 

\noindent
(5) \textbf{PolyPrompt+Expand+XLSum} is the \textit{PolyPrompt} trained on the 7 target datasets, 15 high-resource datasets, and a multilingual summarization dataset (XL-Sum).

\paragraph{Parameters}
The \textit{PolyPrompt} model is built on the mT5 \cite{linting2021mt5} base version \cite{wolf2020transformers} with 580 million parameters.
We used token limits of size $512$ and $64$ for input and output sequences, respectively. All models have a learning rate of $1e-4$, with the batch size set to $18$, and were trained for $20$ epochs. During training, checkpoints were saved every $1,000$ steps. The model with the best performance on the validation set was selected.

\paragraph{Training Data Construction}
Some datasets have a large number of training samples, for example, XNLI has $4.5$ million training samples.
To reduce the expensive computational cost of our experiments, we randomly sampled $3,000$ samples from the training set for each language of the target datasets, and $5,000$ samples from each expanding dataset. These selected samples will serve as the training set for multilingual multitask prompt training with different experiment scenarios.

\paragraph{Experimental Scenario} 
\label{sec:settings}
We consider three experimental scenarios:
(1) \textbf{In-language training}, fine-tuned on golden data in all target languages. Like \citet{pmlr-v119-hu20b}, we use the translations from English released by \citet{pmlr-v119-hu20b} as the golden training samples for the target language for the XQuAD, MLQA, XNLI, and PAWS-X datasets, which have only English training sets. (2) \textbf{Cross-lingual zero-shot transfer} \cite{pmlr-v119-hu20b}, where the model is fine-tuned only on the training set in English. (3) \textbf{Cross-task \& cross-lingual zero-shot transfer}, where a model is evaluated on tasks and languages that did not appear in its training dataset.

\begin{table*}[!ht]
\renewcommand\tabcolsep{2.8pt}
  \centering \footnotesize
    \begin{tabular}{lcccccccccccc}
    \toprule
    & \multicolumn{6}{c}{Question Answering}        & Sentiment & Topic & \multicolumn{2}{c}{Sentence Pair} &  &  \\
\cmidrule{2-11}          & \multicolumn{2}{c}{XQuAD} & \multicolumn{2}{c}{TyDiQA} & \multicolumn{2}{c}{MLQA} & MARC  & MLDOC & PAWS-X & XNLI  & Avg.      &  Sig. \\
\cmidrule{1-11}    Metrics & F1    & EM    & F1    & EM    & F1    & EM    & Acc.  & Acc.  & Acc.  & Acc.  &       &  \\
\midrule
    \multicolumn{12}{l}{In-language training}                                                   &  \\
     \midrule
    Vanilla mT5   & 72.93 & 57.22 & 81.44 & 70.78 & 62.93 & 44.61 & 91.71 & 93.99 & 84.85 & 69.52 & 73.00 & - \\
    \textit{PolyPrompt} & 73.65 & 58.17 & 81.63 & 70.32 & 64.90 & 46.44 & 91.66 & 93.80 & 85.09 & 71.82 & 73.75 & 1.91E-03 \\
    \quad \textit{+Expand} & \textbf{74.15} & \textbf{58.93} & 82.00 & 70.69 & 64.95 & 46.57 & \textbf{91.77} & 93.95 & 84.76 & \textbf{72.28} & 74.00 & 1.54E-03 \\
    \quad \textit{+Expand+XLSum} & 73.35 & 58.01 & 82.37 & \textbf{71.47} & 64.88 & 46.36 & 91.57 & 94.04 & 86.88 & 71.71 & 74.06 & 1.03E-04 \\
    \quad \textit{+Expand+PANX} & 73.73 & 58.43 & \textbf{82.75} & 71.70 & \textbf{65.02} & \textbf{46.60} & 91.55 & \textbf{94.09} & \textbf{87.10} & 72.12 & \textbf{74.31} & 1.03E-04 \\
    \midrule
    \multicolumn{12}{l}{Cross-lingual zero-shot transfer}                                         &  \\
    \midrule
    Vanilla mT5   & 62.49 & 44.51 & 64.67 & 47.46 & 57.16 & 38.92 & 89.75 & 85.74 & 78.24 & 55.54 & 62.45 & - \\
    \textit{PolyPrompt} & 64.01 & 46.33 & 65.47 & 49.57 & 58.19 & 39.92 & 89.85 & 86.01 & 81.10 & 62.95 & 64.34 & 3.96E-05 \\
    \quad \textit{+Expand} & \textbf{65.31} & \textbf{48.07} & \textbf{66.11} & \textbf{50.39} & \textbf{59.48} & \textbf{41.71} & \textbf{90.14} & \textbf{86.84} & \textbf{81.60} & \textbf{64.57} & \textbf{65.42} & 3.96E-05 \\
    \quad \textit{+Expand+XLSum} & 57.50 & 40.56 & 63.45 & 46.94 & 54.97 & 37.51 & 89.60 & 86.43 & 80.35 & 60.93 & 61.82 & \textcolor{gray}{0.18} \\
    \quad \textit{+Expand+PANX} & 64.67 & 47.51 & 65.08 & 48.38 & 59.40 & 41.41 & 89.75 & 86.66 & 81.13 & 63.44 & 64.74 & 8.63E-05 \\
\bottomrule
    \end{tabular}
    \vspace{-7pt}
      \caption{
      Overall results of the models explored in this work on $7$ multilingual datasets from $4$ NLP tasks. Values in \textbf{bold} represent the best performance in a particular setting (e.g. in-language training). \textit{``Avg.''} denotes the average performance of the $7$ datasets, and ``-'' means not applicable. \textit{``Sig.''} is the \textit{``significance test''}, where gray values indicate that the evaluated model failed the significance test ($p>0.05$).
      }
     
  \label{tab:mlpp}%
\end{table*}%

\begin{figure*}[!h]
    \centering
    \includegraphics[width=0.98\linewidth]{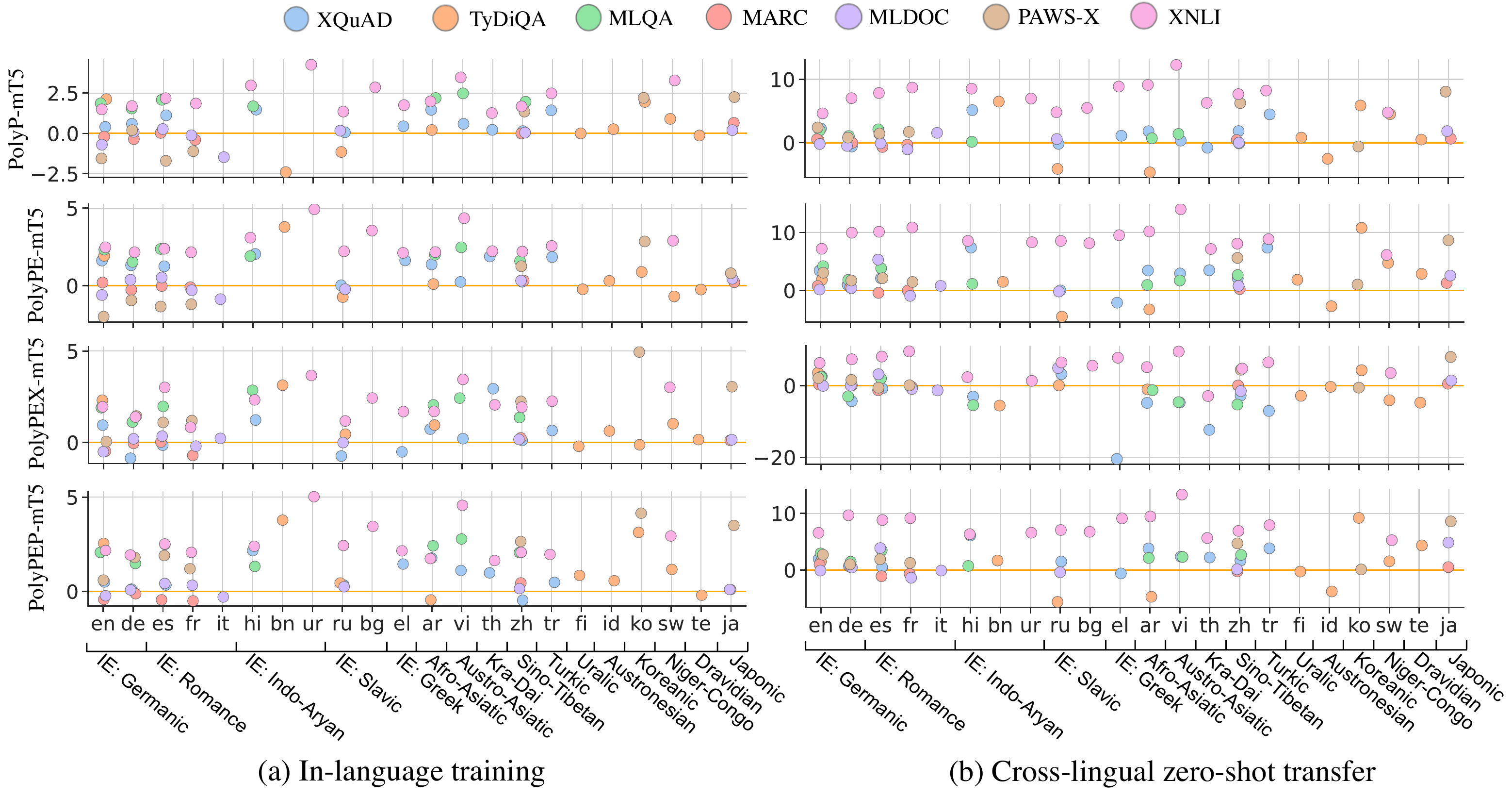}
    \vspace{-6pt}
    \caption{
    The relative performance improvement of \textit{PolyPrompt} and its variants over the \textit{vanilla mT5 (mT5)} at the language-level.
    ``IE'' denotes the ``Indo-European''. \textit{PolyP}, \textit{PolyPE}, \textit{PolyPEX}, and \textit{PolyPEP} are abbreviations for \textit{PolyPrompt}, \textit{PolyPrompt+Expand}, \textit{PolyPrompt+Expand+XLSum}, and \textit{PolyPrompt+Expand+PANX}.
}
	\label{fig:mtl-stl-family}
\end{figure*}

\section{Results \& Analysis}
\label{sec:result}

\subsection{Exp-I: Effect of Multitask Prompt Training}
\label{sec:exp1}
The experiment in this section is designed to answer the research question \textbf{Q1} in Sec.\ref{sec:intro}, namely to investigate whether multilingual multitask prompt training (\textit{PolyPrompt}) can achieve improvement, and whether the performance can be further improved by introducing more high-resource datasets.

\subsubsection{Approach}

\noindent
\textbf{Significance tests: }
To examine whether the \textit{PolyPrompt} and its variants are significantly better than the \textit{vanilla mT5}, we perform the significance test with Wilcoxon’s Signed-rank Test \cite{wilcoxon1970critical} at $p=0.05$. The null hypothesis is that the performance of \textit{PolyPrompt} and its variants is indistinguishable from that of \textit{vanilla mT5}.

\subsubsection{Results} 
We detail main observations in Tab.~\ref{tab:mlpp} and Fig.~\ref{fig:mtl-stl-family}:

\noindent
(1) \textbf{\textit{PolyPrompt} can achieve improvement, especially with the introduction of high-resource datasets.}
Compared to \textit{Vanilla mT5}, the average performance of \textit{PolyPrompt} and its variants (e.g. \textit{PolyPrompt+Expand}) was greatly improved on the 7 datasets of 4 tasks with both the \textit{in-language training} and \textit{cross-lingual zero-shot transfer} settings, other than the  \textit{PolyPrompt+Expand+XLSum} with the \textit{cross-lingual zero-shot transfer} setting ($p=0.18>0.05$).
Furthermore, the best systems for the \textit{in-language training} and \textit{cross-lingual zero-shot transfer} scenarios are \textit{PolyPrompt+Expand+PANX} (5 out of 7) and \textit{PolyPrompt+Expand} (7 out of 7), respectively, illustrating the effectiveness of introducing high-resource expanding datasets.

\begin{table*}[htb!]
  \centering \scriptsize
  \renewcommand\tabcolsep{0.4pt}
    \renewcommand\arraystretch{0.9}  
    \begin{tabular}{cccc cccc cccc cccc cccc cccc cccc}
    \toprule
        \multicolumn{20}{c}{\textbf{PolyPrompt vs. mT5}} &
        \multicolumn{4}{c}{\textbf{PPE vs. mT5}} &
        \multicolumn{4}{c}{\textbf{PPEP vs. mT5}} \\
 \cmidrule(lr){1-20} \cmidrule(lr){21-24} \cmidrule(lr){25-28}
 \multicolumn{4}{c}{\textbf{PAWS-X}} & \multicolumn{4}{c}{\textbf{XNLI}} & \multicolumn{4}{c}{\textbf{XQuAD}} &  \multicolumn{4}{c}{\textbf{MLQA}} & \multicolumn{4}{c}{\textbf{TyDiQA}} & \multicolumn{4}{c}{\textbf{TyDiQA}} & \multicolumn{4}{c}{\textbf{TyDiQA}} \\
 \midrule
 \multicolumn{4}{c}{\textbf{\textcolor{bar_green}{M1}:85.09}} & 
 \multicolumn{4}{c}{\textbf{\textcolor{bar_green}{M1}:71.82}} & 
 \multicolumn{4}{c}{\textbf{\textcolor{bar_green}{M1}:73.65}} & 
 \multicolumn{4}{c}{\textbf{\textcolor{bar_green}{M1}:64.90}} &
 \multicolumn{4}{c}{\textbf{\textcolor{bar_green}{M1}:81.63}} &
  \multicolumn{4}{c}{\textbf{\textcolor{bar_green}{M1}:82.00}} & 
 \multicolumn{4}{c}{\textbf{\textcolor{bar_green}{M1}:82.75}} \\
  \multicolumn{4}{c}{\textbf{\textcolor{bar_purple}{M2}:84.85}} & 
 \multicolumn{4}{c}{\textbf{\textcolor{bar_purple}{M2}:69.52}} & 
 \multicolumn{4}{c}{\textbf{\textcolor{bar_purple}{M2}:72.93}} & 
 \multicolumn{4}{c}{\textbf{\textcolor{bar_purple}{M2}:62.93}} &
 \multicolumn{4}{c}{\textbf{\textcolor{bar_purple}{M2}:81.44}} &
  \multicolumn{4}{c}{\textbf{\textcolor{bar_purple}{M2}:81.44}} & 
 \multicolumn{4}{c}{\textbf{\textcolor{bar_purple}{M2}:81.44}} \\
 \midrule
 \multicolumn{4}{c}{\multirow{5}[2]{*}{\includegraphics[scale=0.15]{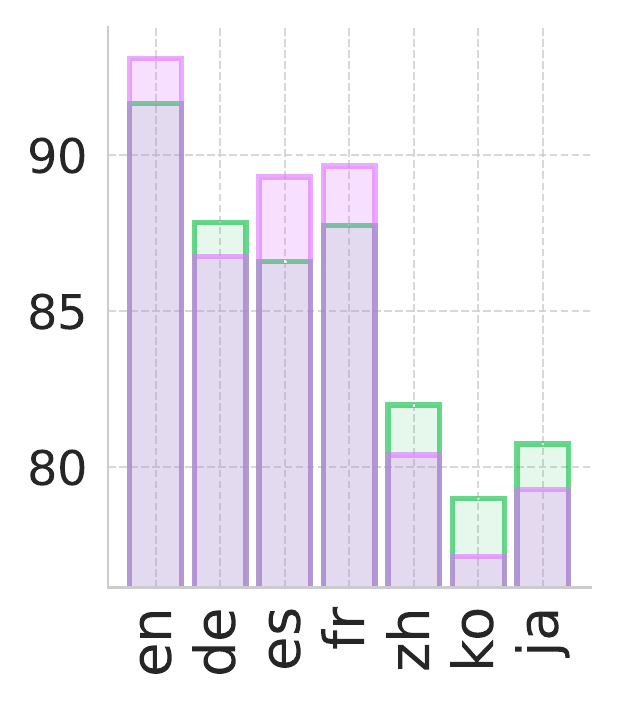}}} & 
\multicolumn{4}{c}{\multirow{5}[2]{*}{\includegraphics[scale=0.15]{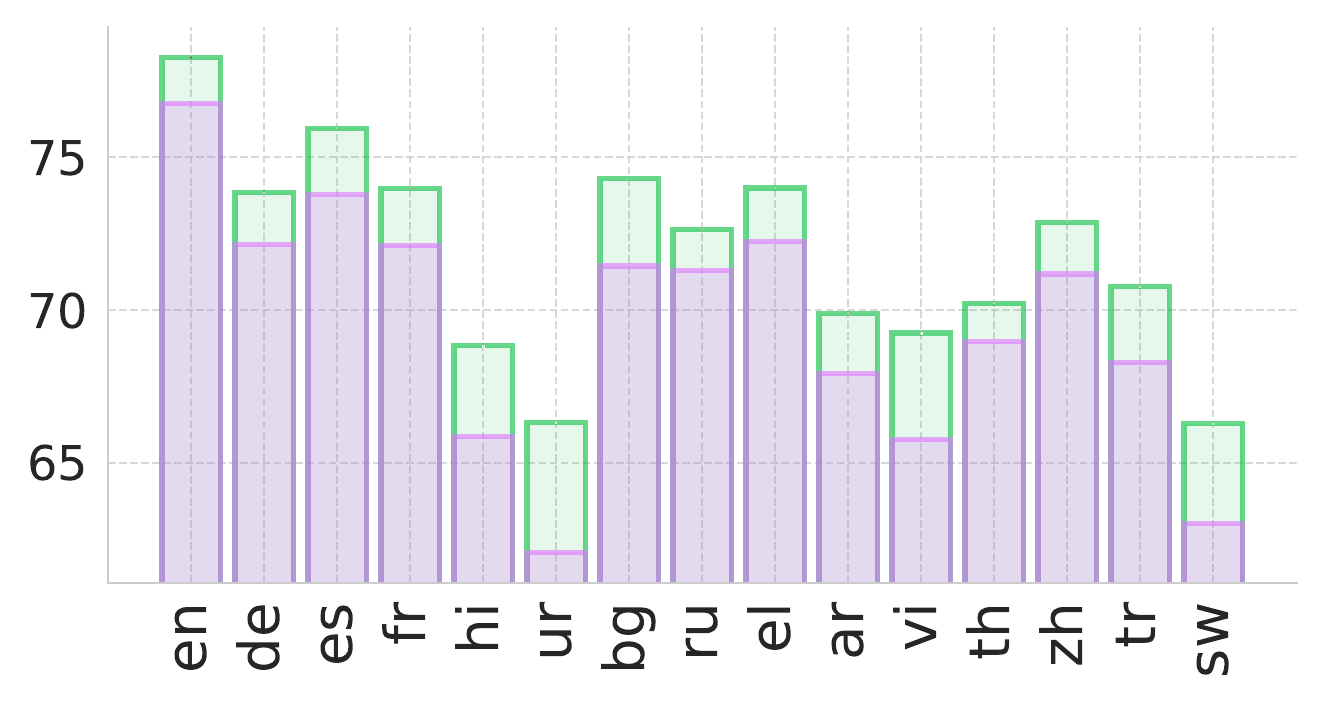}}} &
\multicolumn{4}{c}{\multirow{5}[2]{*}{\includegraphics[scale=0.15]{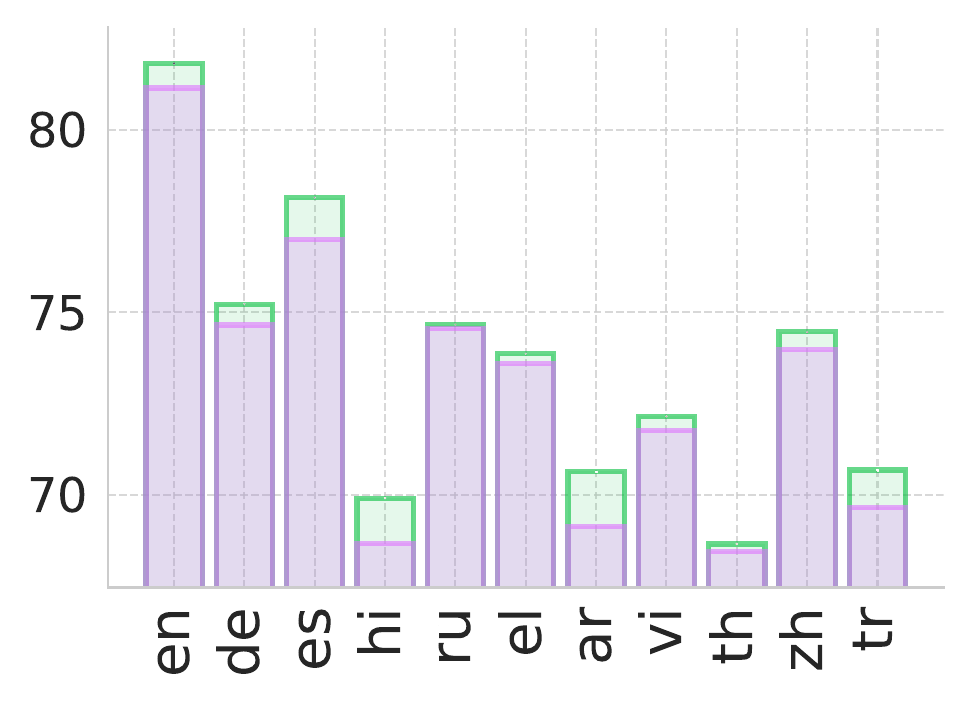}}} & 
\multicolumn{4}{c}{\multirow{5}[2]{*}{\includegraphics[scale=0.15]{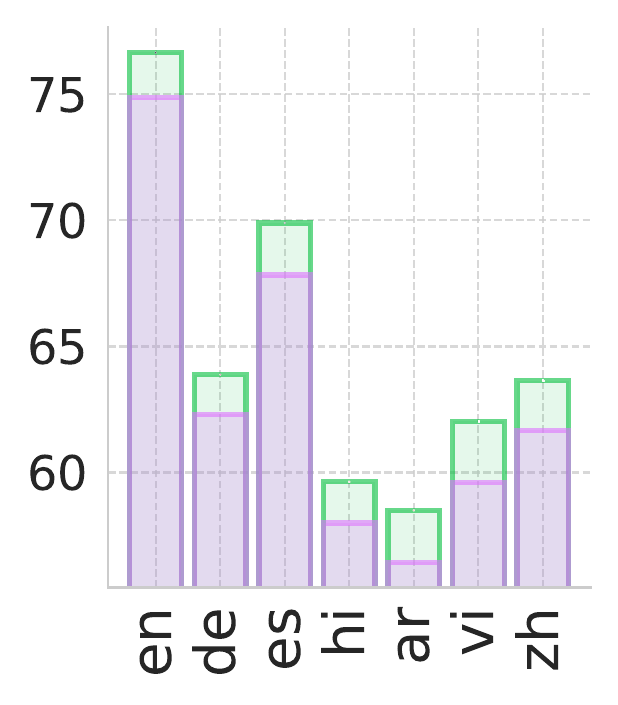}}} &
\multicolumn{4}{c}{\multirow{5}[2]{*}{\includegraphics[scale=0.15]{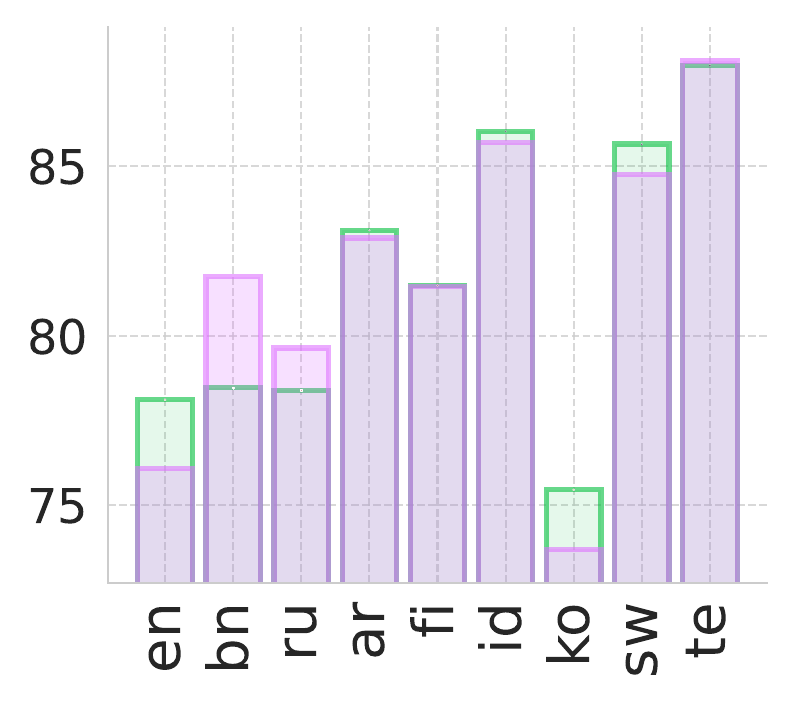}}} & 
\multicolumn{4}{c}{\multirow{5}[2]{*}{\includegraphics[scale=0.15]{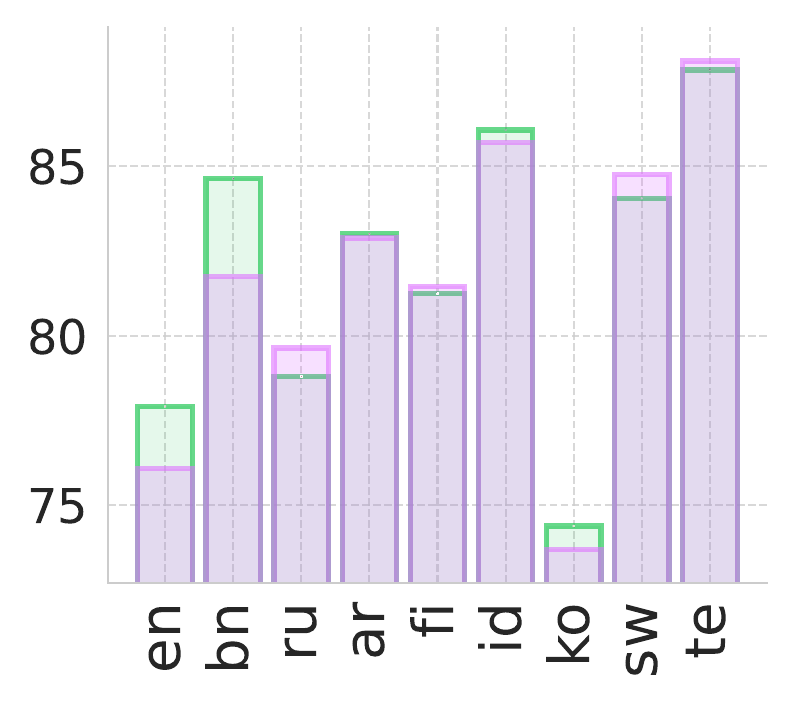}}} &
\multicolumn{4}{c}{\multirow{5}[2]{*}{\includegraphics[scale=0.15]{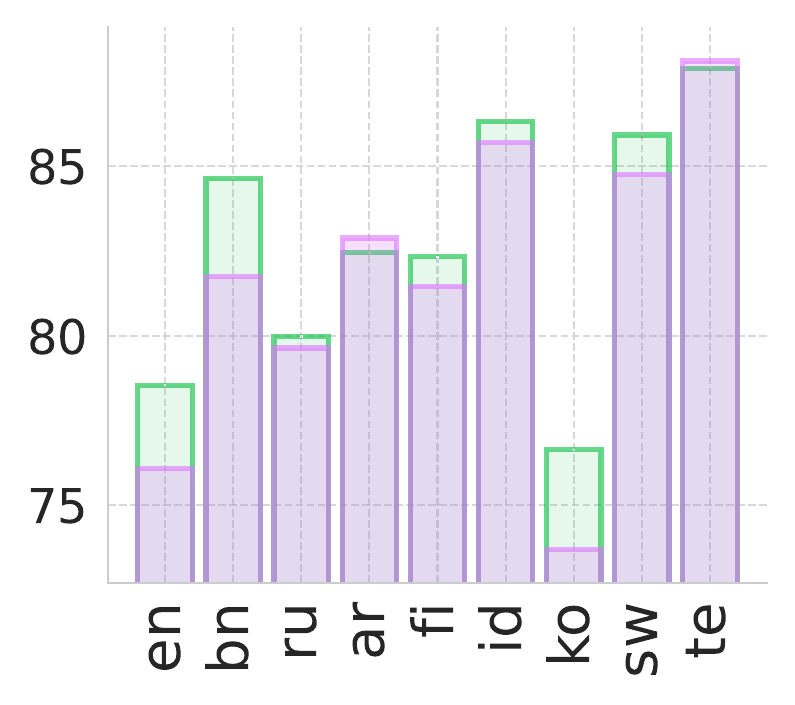}}} \\ \\ \\ \\ \\ \\ \\
    \midrule
        \multicolumn{4}{c}{\texttt{t1Len}} & \multicolumn{4}{c}{\texttt{t1Len}} & \multicolumn{4}{c}{\texttt{cLen}} & \multicolumn{4}{c}{\texttt{cLen}} & \multicolumn{4}{c}{\texttt{cLen}}  & \multicolumn{4}{c}{\texttt{cLen}} & \multicolumn{4}{c}{\texttt{cLen}} \\
    \midrule
        \multicolumn{4}{c}{\multirow{5}[2]{*}{\includegraphics[scale=0.15]{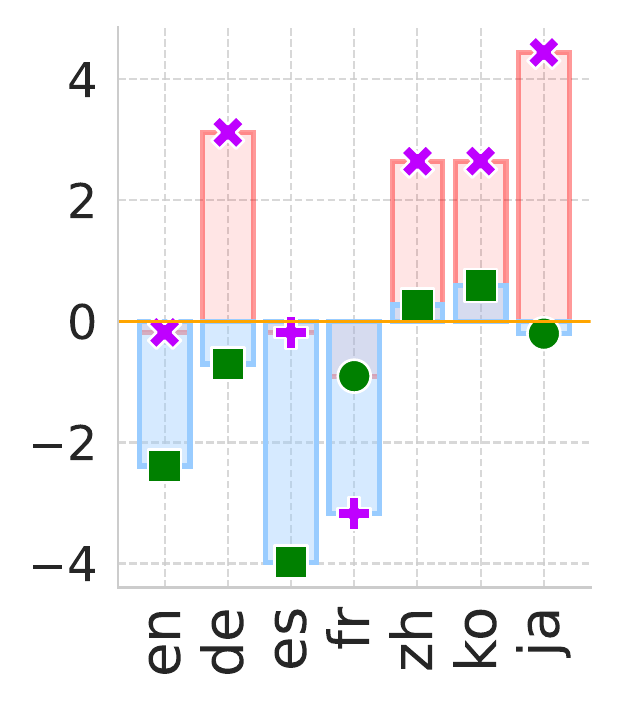}}} & 
        \multicolumn{4}{c}{\multirow{5}[2]{*}{\includegraphics[scale=0.15]{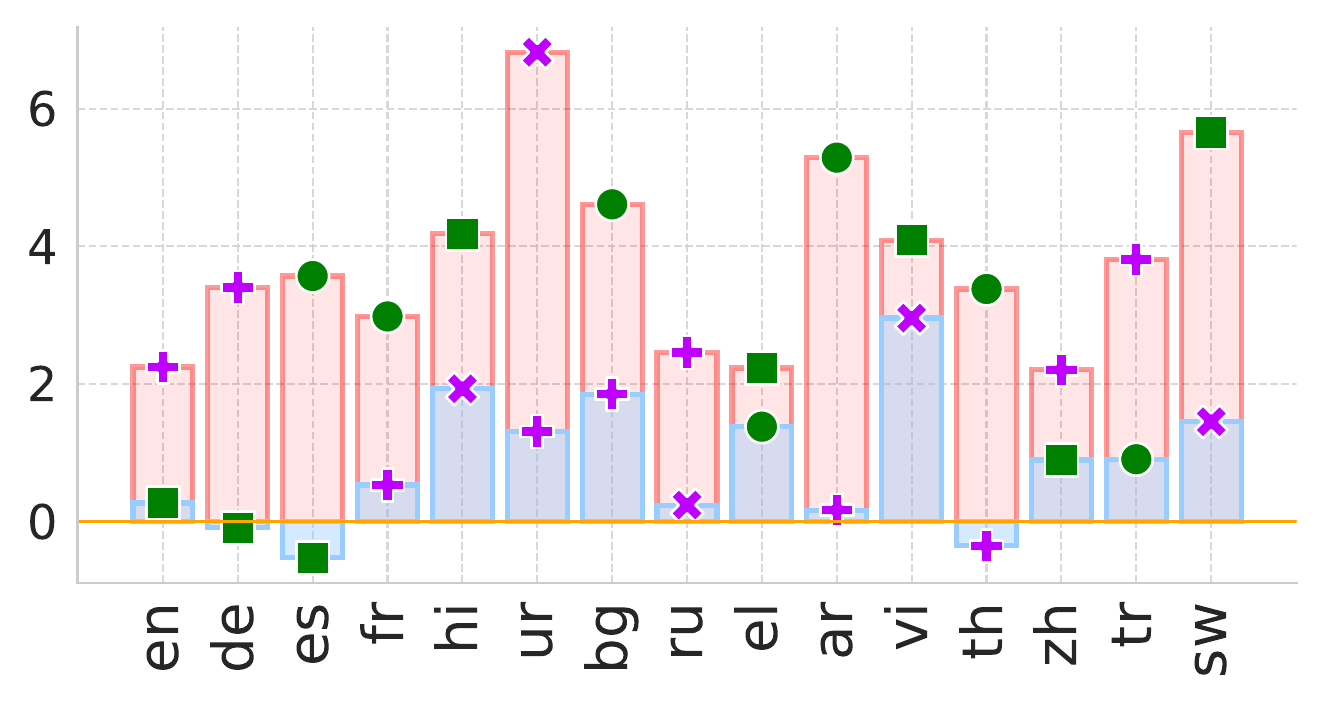}}} &
         \multicolumn{4}{c}{\multirow{5}[2]{*}{\includegraphics[scale=0.15]{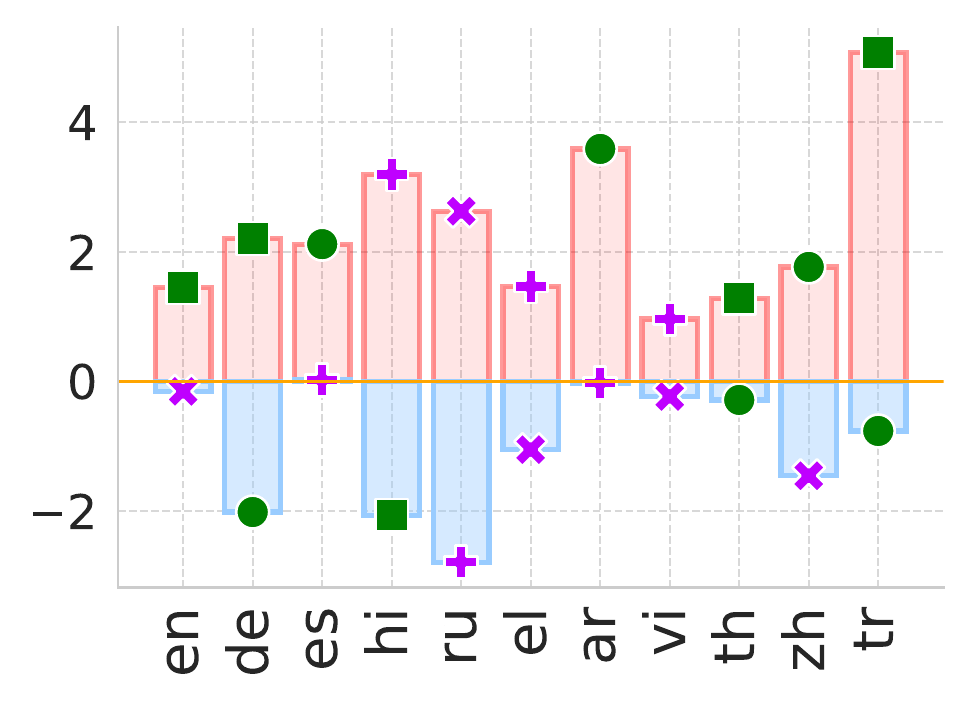}}} & 
\multicolumn{4}{c}{\multirow{5}[2]{*}{\includegraphics[scale=0.15]{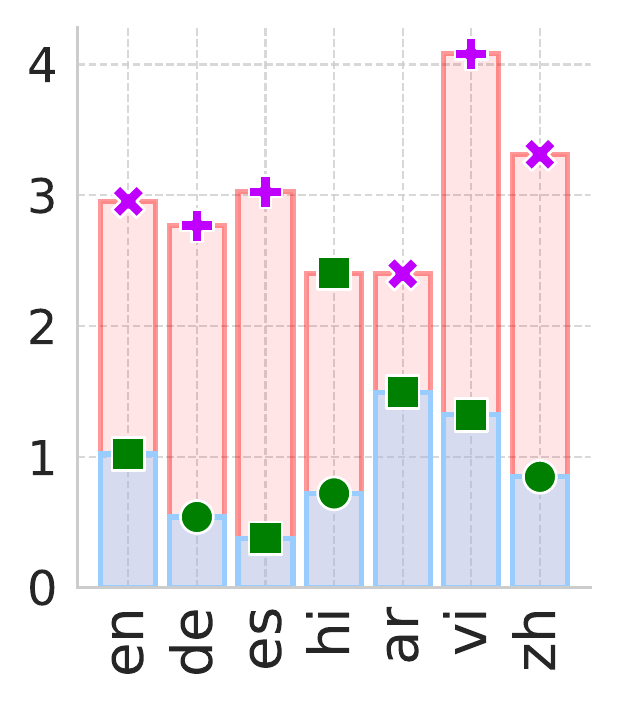}}}  &
\multicolumn{4}{c}{\multirow{5}[2]{*}{\includegraphics[scale=0.15]{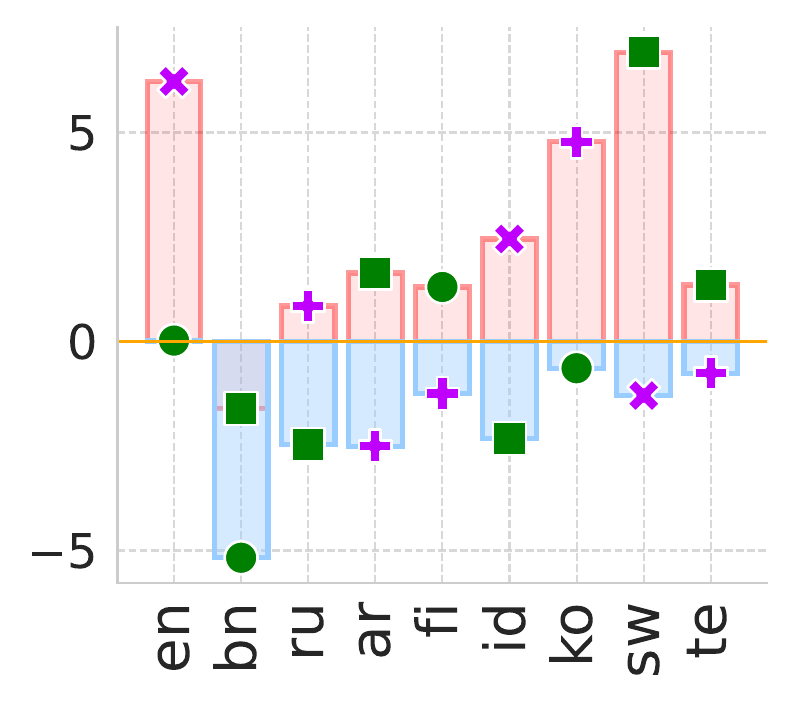}}} &
\multicolumn{4}{c}{\multirow{5}[2]{*}{\includegraphics[scale=0.15]{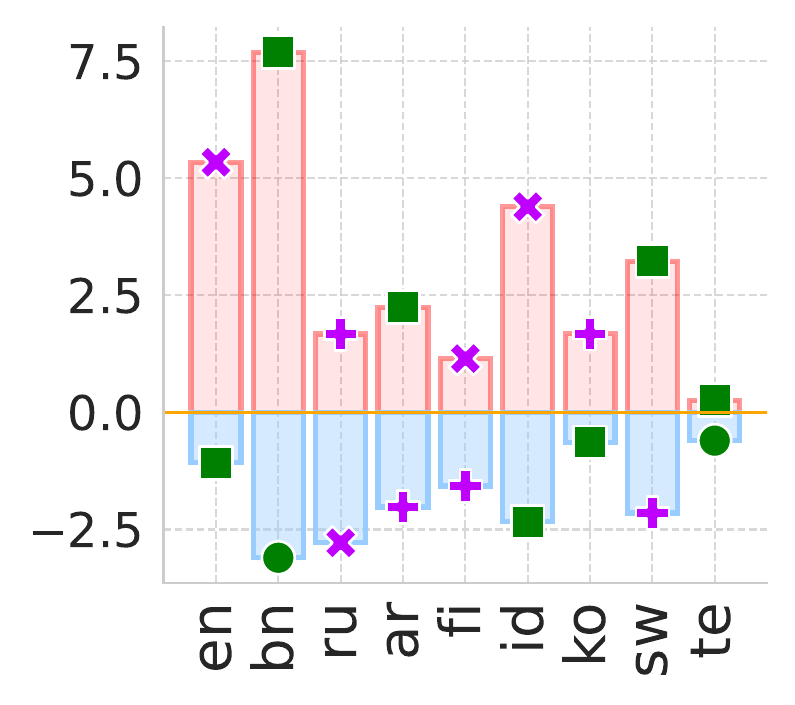}}} &
\multicolumn{4}{c}{\multirow{5}[2]{*}{\includegraphics[scale=0.15]{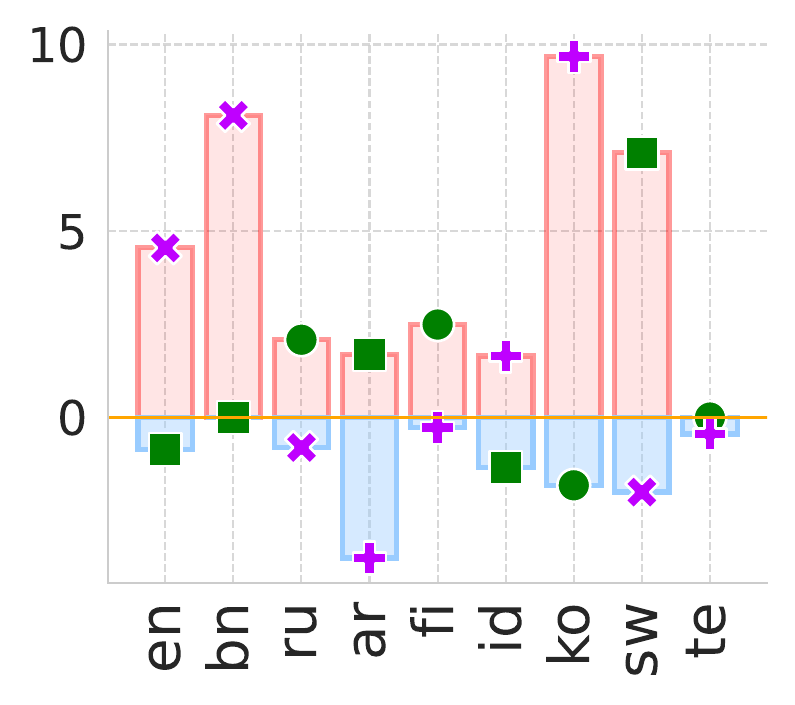}}} \\ \\ \\ \\ \\ \\ \\
\midrule
        \multicolumn{4}{c}{\texttt{t2Len}} & \multicolumn{4}{c}{\texttt{t2Len}} & \multicolumn{4}{c}{\texttt{qLen}} & \multicolumn{4}{c}{\texttt{qLen}} & \multicolumn{4}{c}{\texttt{qLen}} & \multicolumn{4}{c}{\texttt{qLen}} & \multicolumn{4}{c}{\texttt{qLen}} \\
\midrule
\multicolumn{4}{c}{\multirow{5}[2]{*}{\includegraphics[scale=0.15]{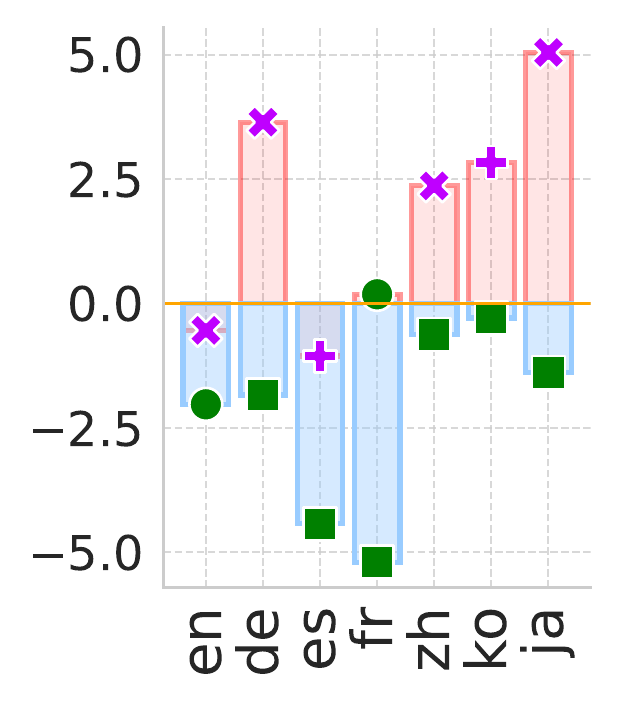}}} & 
\multicolumn{4}{c}{\multirow{5}[2]{*}{\includegraphics[scale=0.15]{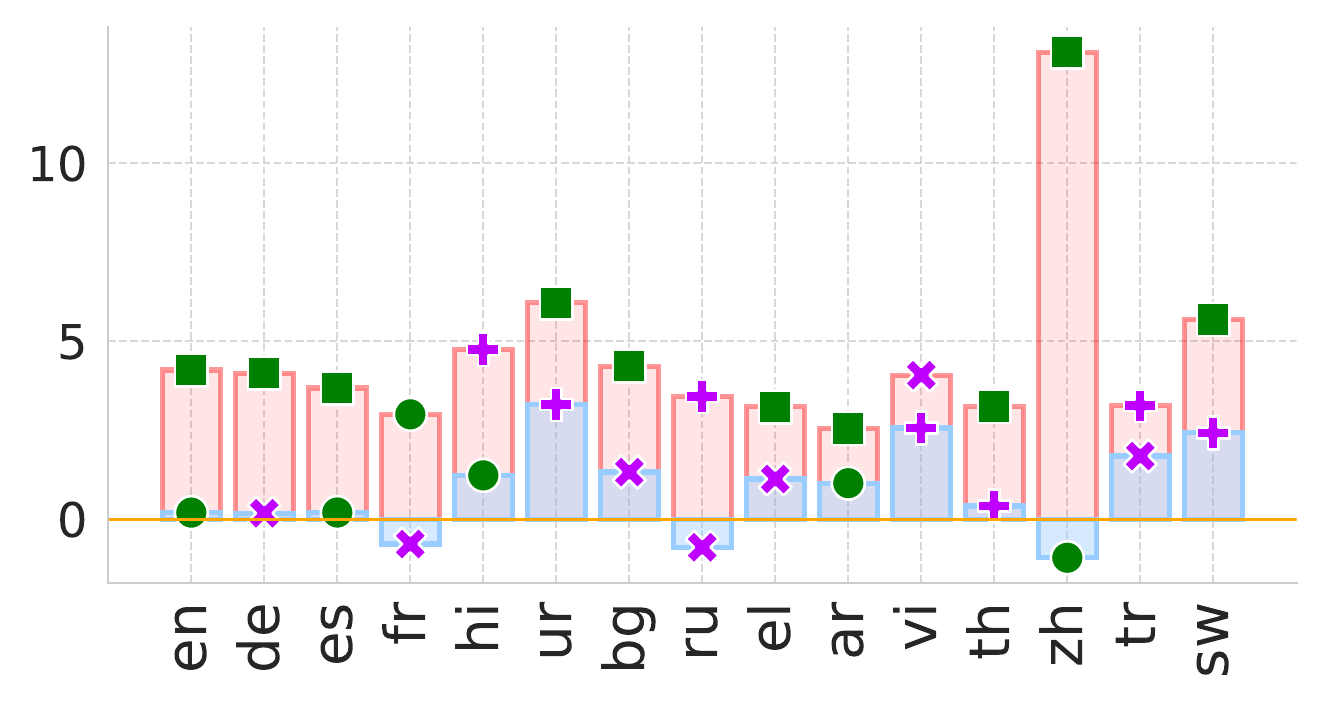}}} &
         \multicolumn{4}{c}{\multirow{5}[2]{*}{\includegraphics[scale=0.15]{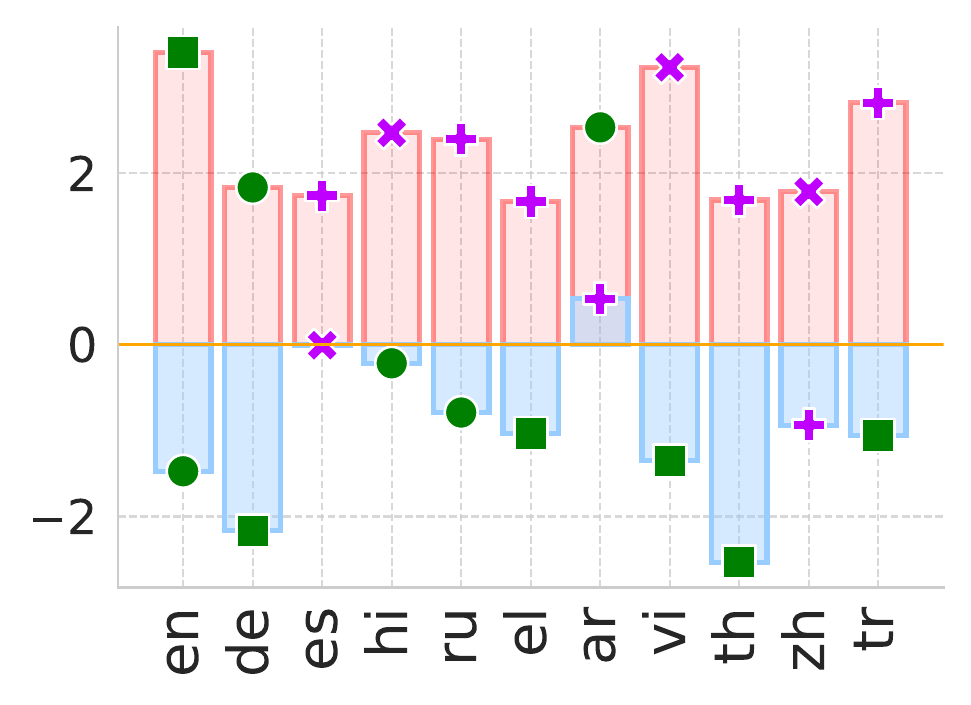}}} & 
\multicolumn{4}{c}{\multirow{5}[2]{*}{\includegraphics[scale=0.15]{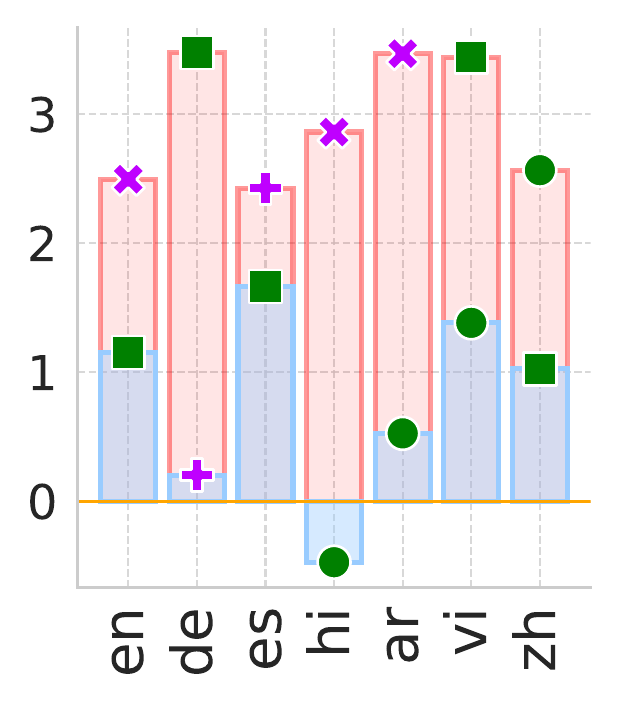}}}  &
\multicolumn{4}{c}{\multirow{5}[2]{*}{\includegraphics[scale=0.15]{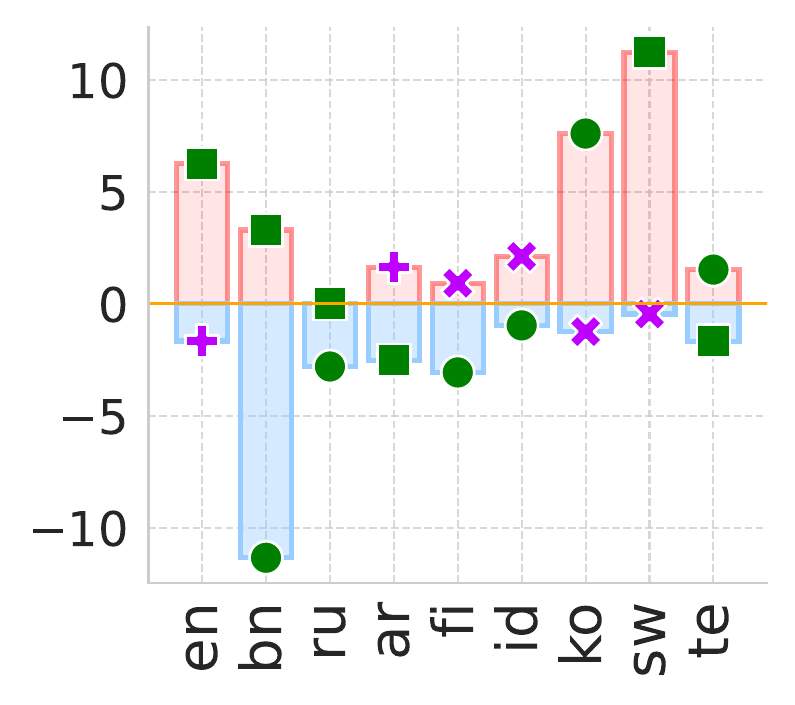}}} &
\multicolumn{4}{c}{\multirow{5}[2]{*}{\includegraphics[scale=0.15]{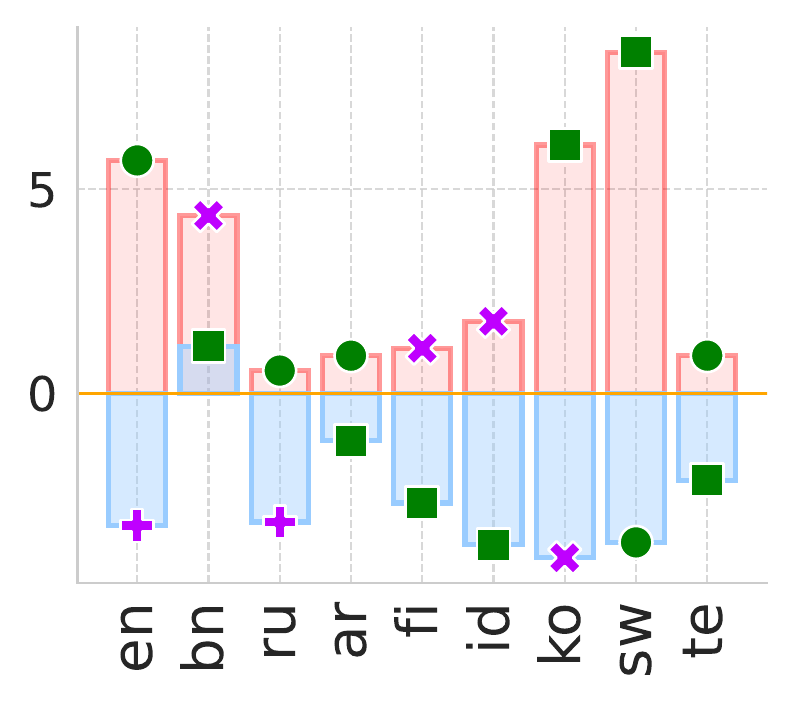}}} &
\multicolumn{4}{c}{\multirow{5}[2]{*}{\includegraphics[scale=0.15]{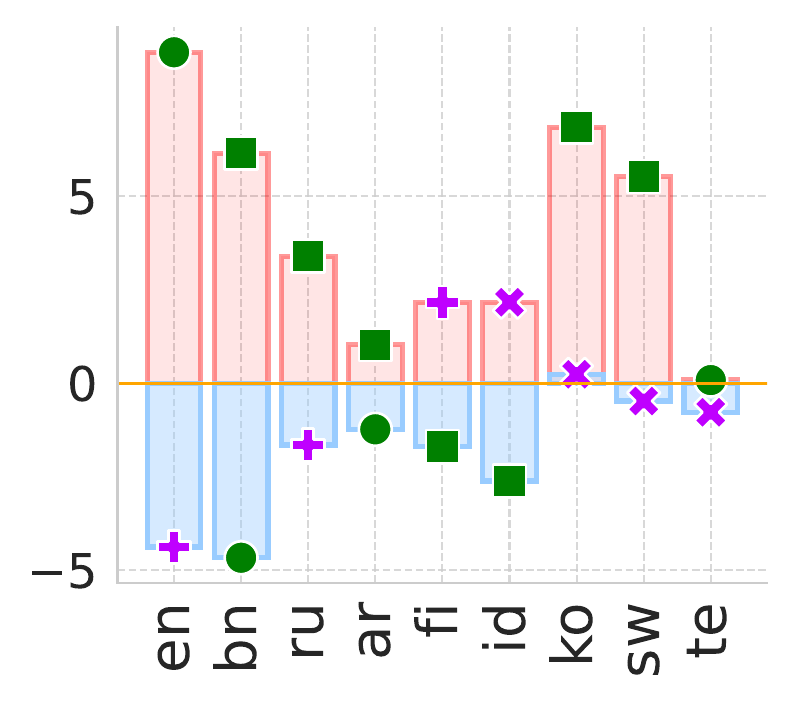}}} \\ \\ \\ \\ \\ \\ \\
    \bottomrule 
    \end{tabular}
    \vspace{-7pt}
      \caption{
      The pairwise model diagnosis of \textit{PolyPrompt} (and its variants) and \textit{vanilla mT5 (mT5)}.
      The model name on the left side of ``vs.'' denotes M1, and the right side represents M2.
      The bar charts in the first row are the overall performance of \textcolor{bar_green}{M1} (green bars) and \textcolor{bar_purple}{M2} (purple bars) across different languages.
      The bar charts after the second row represent the relative performance improvement of \textit{PolyPrompt} and its variants over \textit{vanilla mT5}, where the heights of the red and blue bars represent the maximum positive and maximum negative gains.
      \textcolor{icon_purple}{\textbf{$\mathbf{\times}$}}, 
      \textcolor{icon_purple}{\textbf{+}}, 
      \textcolor{icon_green}{\faCircle}, and 
      \textcolor{icon_green}{$\medblacksquare$} denotes the ``extra-small (XS)'',``small (S)'',``large (L)'', and ``extra-large (XL)'', respectively.
      \textit{PPE} and \textit{PPEP} denote \textit{PolyPrompt+Expand} and \textit{PolyPrompt+Expand+PANX}, respectively.
      }
  \label{tab:pp-mt5-bar}%
\end{table*}%

\begin{figure}[h!]
  \centering \scriptsize
  \renewcommand\tabcolsep{0.3pt}
    \renewcommand\arraystretch{1.1}  
    \begin{tabular}{cccc cccc cccc cccc cccccc}
    \toprule
 \multicolumn{8}{c}{\textbf{PAWS-X}} & &&&&&&
 \multicolumn{8}{c}{\textbf{XNLI}}  \\
\cmidrule(lr){1-8}\cmidrule(lr){15-22}
    \multicolumn{4}{c}{\texttt{t1Len}} & \multicolumn{4}{c}{\texttt{t2Len}} & &&&&&&
    \multicolumn{4}{c}{\texttt{t1Len}} & \multicolumn{4}{c}{\texttt{t2Len}}  \\
\cmidrule(lr){1-8}\cmidrule(lr){15-22}
    \multicolumn{4}{c}{\multirow{5}[2]{*}{\includegraphics[scale=0.13]{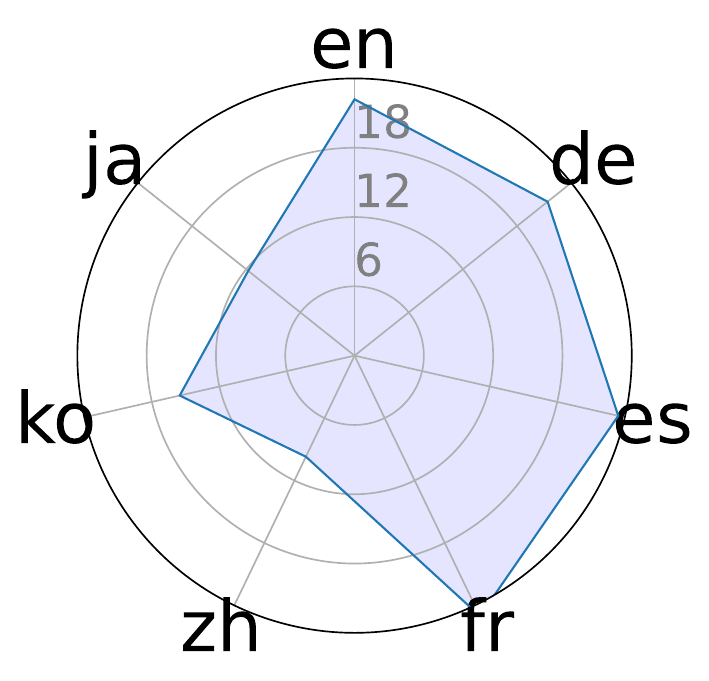}}} & 
    \multicolumn{4}{c}{\multirow{5}[2]{*}{\includegraphics[scale=0.13]{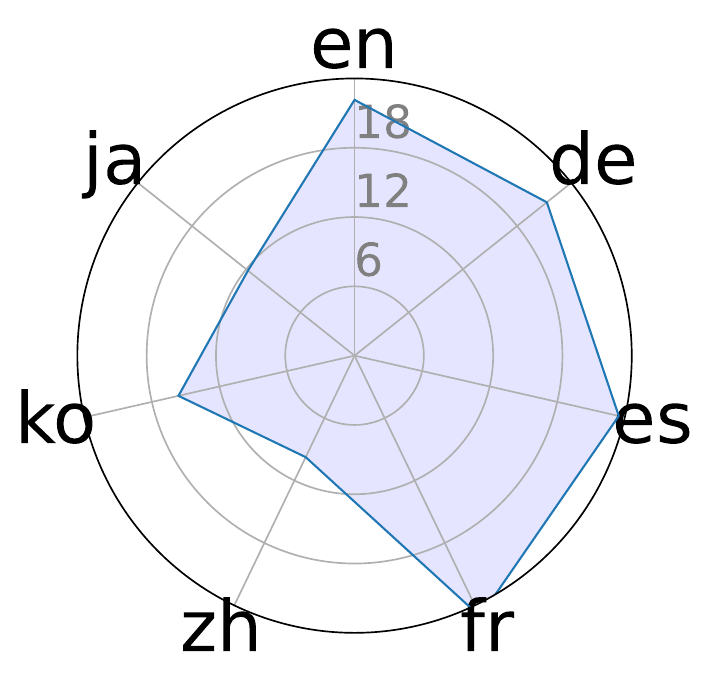}}} & &&&&&&
    \multicolumn{4}{c}{\multirow{5}[2]{*}{\includegraphics[scale=0.13]{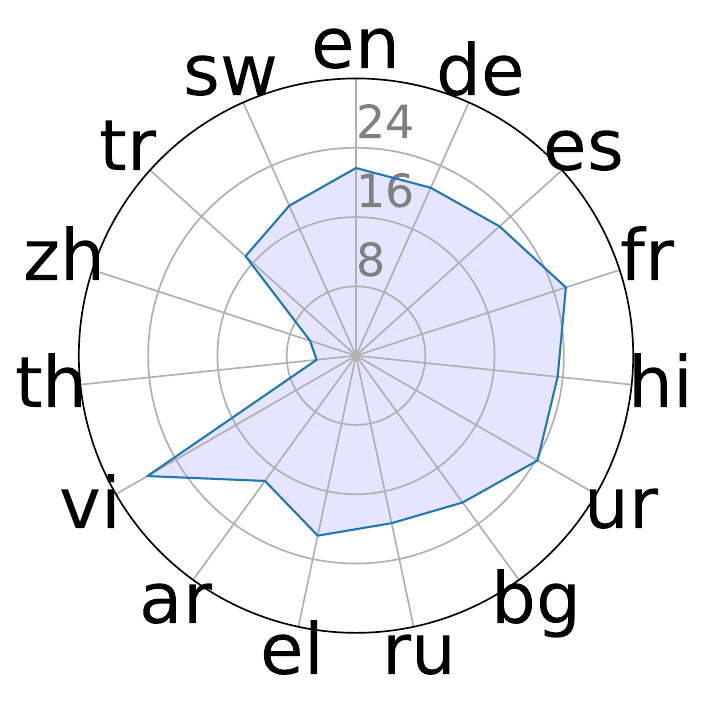}}} &
    \multicolumn{4}{c}{\multirow{5}[2]{*}{\includegraphics[scale=0.13]{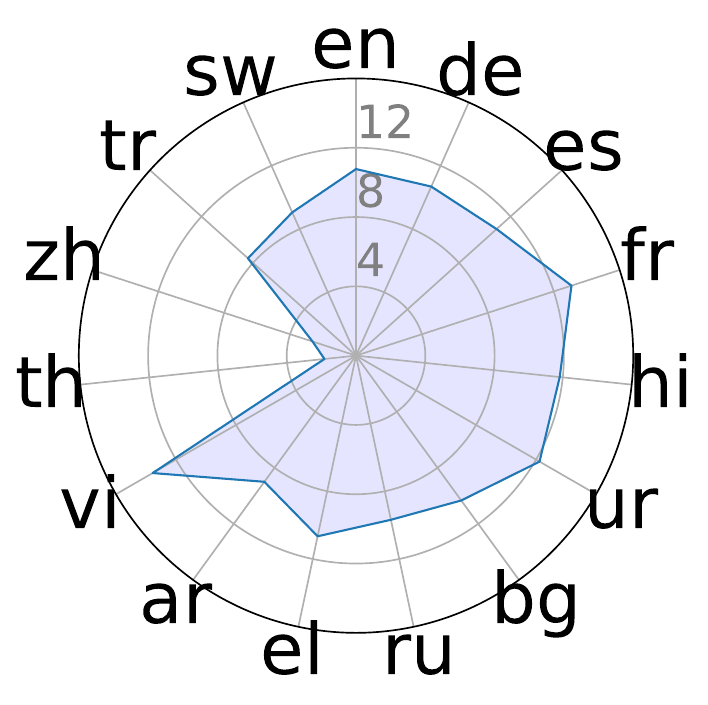}}} 
    \\ \\ \\ \\ \\ 
    \bottomrule 
    \end{tabular}
    \vspace{-7pt}
      \caption{Dataset bias of \texttt{PAWS-X} and \texttt{XNLI} characterized by $\phi_{p}$ defined in Sec.~\ref{sec:exp2-approach}. 
      }
  \label{fig:data-bias}%
\end{figure}%

\noindent
(2) \textbf{It is more beneficial for \textit{PolyPrompt} to introduce high-resource English datasets in the cross-language zero-shot transfer.}
In Fig.~\ref{fig:mtl-stl-family}-(b), we observe that \textit{PolyPrompt} achieved performance gains in many languages belonging to different datasets in the \textit{cross-language zero-shot transfer} scenarios. 
When the external English datasets (\textit{PolyPrompt+Expand}) are introduced, more languages gained performance improvements. 
However, when multilingual datasets (e.g. PANX and XLSum) were introduced (\textit{PolyPrompt+Expand+PANX} and \textit{PolyPrompt+Expand+XLSum}), the overall performance dropped (observed from Tab.~\ref{tab:mlpp}) and there were more languages with negative relative performance gains (compared to \textit{PolyPrompt+Expand}).

\subsection{Exp-II: Multilingual Interpretable  Evaluation}
\label{sec:interpret}

This section aims at the research question \textbf{Q2} (\textit{How do different characteristics of datasets and languages affect the performance of \textit{PolyPrompt}?}) by introducing a multilingual interpretable evaluation.

\subsubsection{Approach}
\label{sec:exp2-approach}

Interpretable evaluation \cite{liu2021explainaboard,fu2021interpteval,ruder2021xtreme} aims to breakdown the holistic performance (e.g., F1) to a more fine-grained level based on predefined features (e.g., text length) to interpret the model's behavior better.

Below, we list some features of each task explored in this work.
\text{XQuAD, TyDiQA, MLQA}: the length of context (\texttt{cLen}), question (\texttt{qLen}), and answer (\texttt{aLen}). The BLUE score of the answer and context (\texttt{BLUE\_AC}). \text{PAWS-X, XNLI}: the length of $\text{sentence}_{1}$ (\texttt{t1Len}) and $\text{sentence}_{2}$ (\texttt{t2Len}). The BLUE score of the sentence pair (\texttt{BLUE\_t1t2}).
Further detailed interpretable evaluation definition can be found in App.~\ref{sec:interp-define-app}.
We also measure the \textbf{dataset-level} feature $\phi_{p}$, the average feature value over a dataset. For example, $\phi_{\text{aLen}(\text{MLQA})}$ denotes the average answer length of MLQA dataset. Further details can be found in App.~\ref{sec:data-feat-define-app}

\subsubsection{Results} 

Here are the main observations in Tab.~\ref{tab:pp-mt5-bar} and Fig.~\ref{fig:data-bias}:

\noindent
(1) \textbf{Dataset Perspective:} the strengths of \textit{PolyPrompt}  in the co-occurring languages on different datasets are inconsistent due to dataset bias. For example,
\texttt{en}, \texttt{de}, and \texttt{fr} co-occur on PAWS-X and XNLI.
\textit{PolyPrompt} was better in the short sentence$_2$ (\texttt{t2Len:XS}) in PAWS-X, while excelling in the long sentence$_2$ (\texttt{t2Len:XL}) of XNLI. 
This inconsistency results from the dataset bias shown in Fig.~\ref{fig:data-bias}: the
$\phi_{\text{t2Len}}(\text{XNLI-[en,de,fr]})$ < 12 while $\phi_{\text{t2Len}}(\text{PAWS-X-[en,de,fr]})$ > 20.
Therefore, $\phi_{\text{t2Len}}(\text{PAWS-X-[en,de,fr]})$ on bucket \texttt{\text{t2Len}:XS} was close to $\phi_{\text{t2Len}}(\text{XNLI-[en,de,fr]})$ on bucket \texttt{\text{t2Len}:XL}.

\noindent
(2) \textbf{Model Perspective: } 
\textit{PolyPrompt} achieves overall performance improvements on the 7 target datasets, but it cannot perform well on all samples (e.g., worse performance on long sentences).
\textit{PolyPrompt} is better at short context
samples for \text{MLQA} (\texttt{cLen:XS/S}), long context samples for \text{XQuAD} (\texttt{cLen:XL/L}), long $\text{sentence}_2$ for \text{XNLI} (\texttt{t2Len:XL/L}), which is valid for most languages. 
\textbf{Disadvantage analysis: }
 \textit{PolyPrompt} is worse at handling long question samples (\texttt{qLen:XL/L}) for \text{XQuAD}, \text{TyDiQA}, and \text{MLQA}, long $\text{sentence}_1$ samples for XNLI-es, and long $\text{sentence}_1$ and $\text{sentence}_2$ samples in \texttt{zh,ko,ja} of \text{PAWS-X}.

\noindent
(3) \textbf{Language Perspective:} it is difficult for \textit{PolyPrompt} to bring gains for languages that appear only once in the 7 target datasets unless high-resource datasets are introduced.
For example, \textit{PolyPrompt} showed a slight performance improvement over \textit{vanilla mT5} in languages \texttt{bn}, \texttt{fi}, \texttt{id}, and \texttt{te} that only appeared in the TyDiQA dataset. 
When introducing high-resource English datasets, the performance of \texttt{bn} is significantly improved especially for long context and short answers samples, while the performance improvement of \texttt{fi}, \texttt{id}, and \texttt{te} is still limited until a high-resource multilingual training dataset PANX is introduced.
The reason may be that most of the languages in the 7 tasks belong to the IE language family (findings from Fig.~\ref{fig:mtl-stl-family}), and so does the \texttt{bn} language. Therefore, compared to \texttt{fi}, \texttt{id}, and \texttt{te}, it is easier for \texttt{bn}  to get knowledge from neighbor languages in multitask training.

\begin{table*}[htb]
  \centering \scriptsize
    \renewcommand\tabcolsep{2.8pt}
    \begin{tabular}{lllm{10.8cm}}
    \toprule
    \multicolumn{2}{c}{\textbf{Prompt Design}} & \multicolumn{1}{c}{\textbf{Dataset}} & \multicolumn{1}{c}{\textbf{Prompt Template}} \\
    \cmidrule(lr){1-2}\cmidrule(lr){3-3}\cmidrule(lr){4-4}
 \multirow{2}[2]{*}{\makecell{Language \\ Choice}} & In-lingual (zh)  & TyDiQA    & 根据段落的内容回答问题。 | 问题：\texttt{[Q-zh]} | 段落：\texttt{[C-zh]} \\
    \cmidrule(lr){2-4}
    & Cross-lingual (zh)  & TyDiQA    & Answer the question based on the content of the paragraph. | \texttt{[Q-zh]} | Paragraph: \texttt{[C-zh]}  \\
    \midrule
  \multirow{8}[8]{*}{\makecell{Uniformity \\ of \\ Templates}} & \multirow{4}[2]{*}{Unified } & XQuAD & \textcolor{bittersweet}{Answer the question based on the paragraph.} \textcolor{bittersweet}{| Question:} \texttt{[Q-xx]} \textcolor{bittersweet}{| Paragraph:} \texttt{[C-xx]} \\
      &   & MLQA  & \textcolor{bittersweet}{Answer the question based on the paragraph.} \textcolor{bittersweet}{| Question}: \texttt{[Q-xx]} \textcolor{bittersweet}{| Paragraph:} \texttt{[C-xx]} \\
      &  & XNLI  & \textcolor{bittersweet}{Answer the question based on paragraph} 1 and \textcolor{bittersweet}{paragraph} 2 . \textcolor{bittersweet}{| Question :} Do \textcolor{bittersweet}{Paragraph} 1 and \textcolor{bittersweet}{Paragraph} 2 mean the same thing ? | ( A ) Yes . ( B ) No . ( C ) Maybe . \textcolor{bittersweet}{| Paragraph} 1 : \texttt{[T1-xx]} \textcolor{bittersweet}{Paragraph} 2 : \texttt{[T2-xx]}  \\
      &  & MARC  & \textcolor{bittersweet}{Answer the question based on the} review. \textcolor{bittersweet}{| Question:} Can we conclude that the buyer is satisfied with the product based on his review? ｜ (A) Yes. (B) No. | Review: \texttt{[T1-xx]} \\
    \cmidrule(lr){2-4}
  & \multirow{4}[2]{*}{Diversified} & XQuAD  & I have always wondered: \texttt{[Q-xx]} | I searched Wikipedia and this is what I found. What's the answer? | \texttt{[C-xx]} \\
        &  & MLQA  & Context: \texttt{[C-xx]} | I am trying to figure out the answer to the question from the above context. Can you tell me the answer? | Question: \texttt{[Q-xx]} Answer: \\
        &  & XNLI  & Given that \texttt{[T1-xx]} Therefore, it must be true that \texttt{[T2-xx]}? Yes, no, or maybe? \\
        &   & MARC  & I am reading a review that says \texttt{[T1-xx]}. Do you think the review is positive or negative? \\
    \bottomrule
    \end{tabular}
    \vspace{-6pt}
      \caption{Examples of prompt design in this work. ``\texttt{Q}'', ``\texttt{C}'', ``\texttt{T1}'', and ``\texttt{T2}'' denotes the placeholders for question, context, $\text{sentence}_1$, and $\text{sentence}_2$ field, respectively.
      The format ``[text field - language]'' is used to represent text in a specific language, such as ``[Q-zh]'' denotes a question text in Chinese (zh).
     ``\texttt{xx}'' denotes any language, and ``|'' represents the separator of the input field (e.g., question and context). 
      The tokens in \textcolor{bittersweet}{pink} are the co-occurring words in most templates.
      } 
  \label{tab:prompt-type}
\end{table*}

\begin{figure*}[!ht]
	\centering
    \subfigure[$\text{Cross-lingual (CL)} - \text{In-lingual (IL)}$]
	{
	\begin{minipage}[b]{0.43\textwidth}
		\includegraphics[width=1\textwidth]{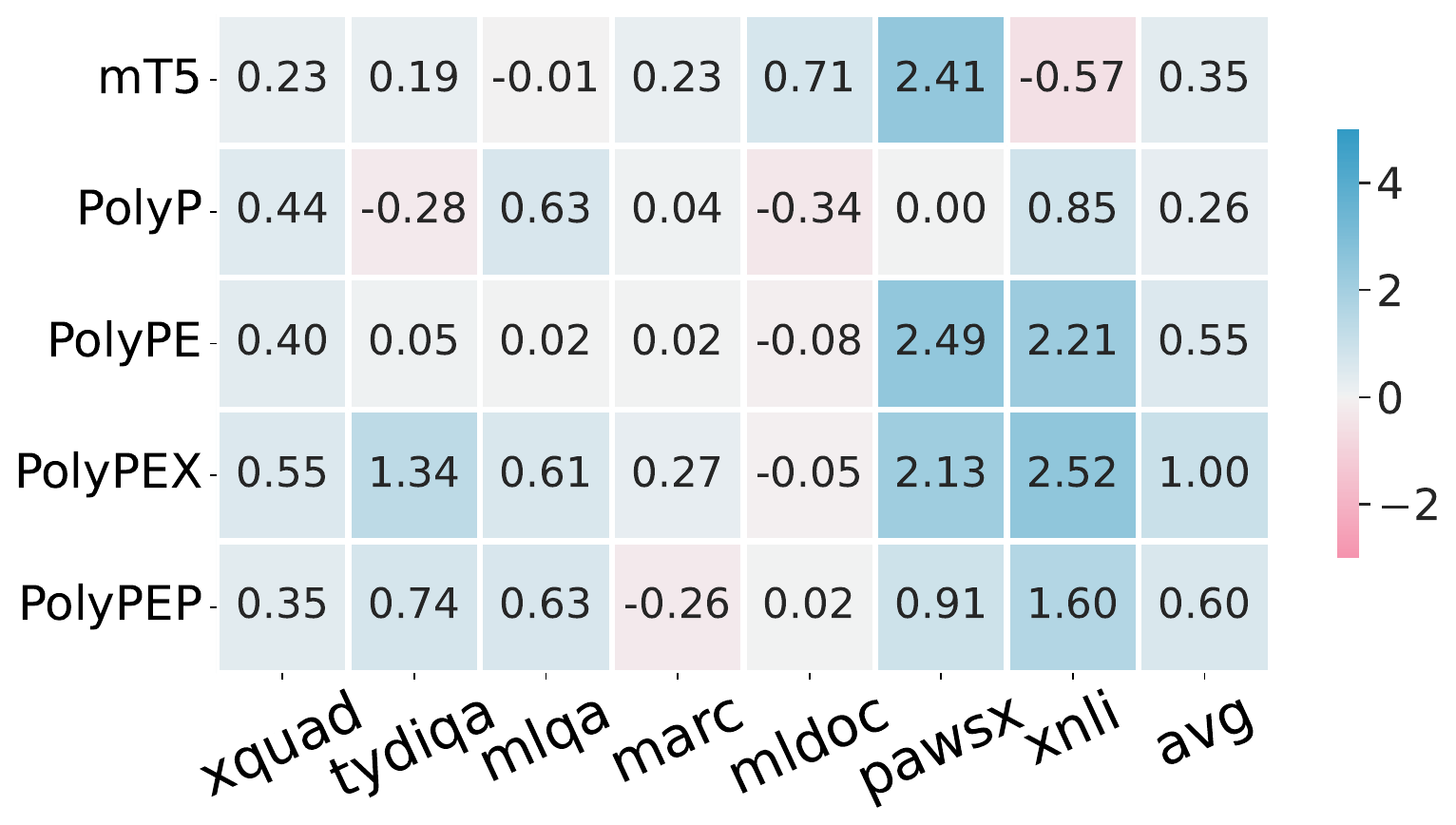}
	\end{minipage}
	} 
	\hspace{-10pt}
	\subfigure[$\textit{PolyPrompt} - \textit{PolyPrompt-v(x)}$]{
	\begin{minipage}[b]{0.43 \textwidth}
        \includegraphics[width=1\textwidth]{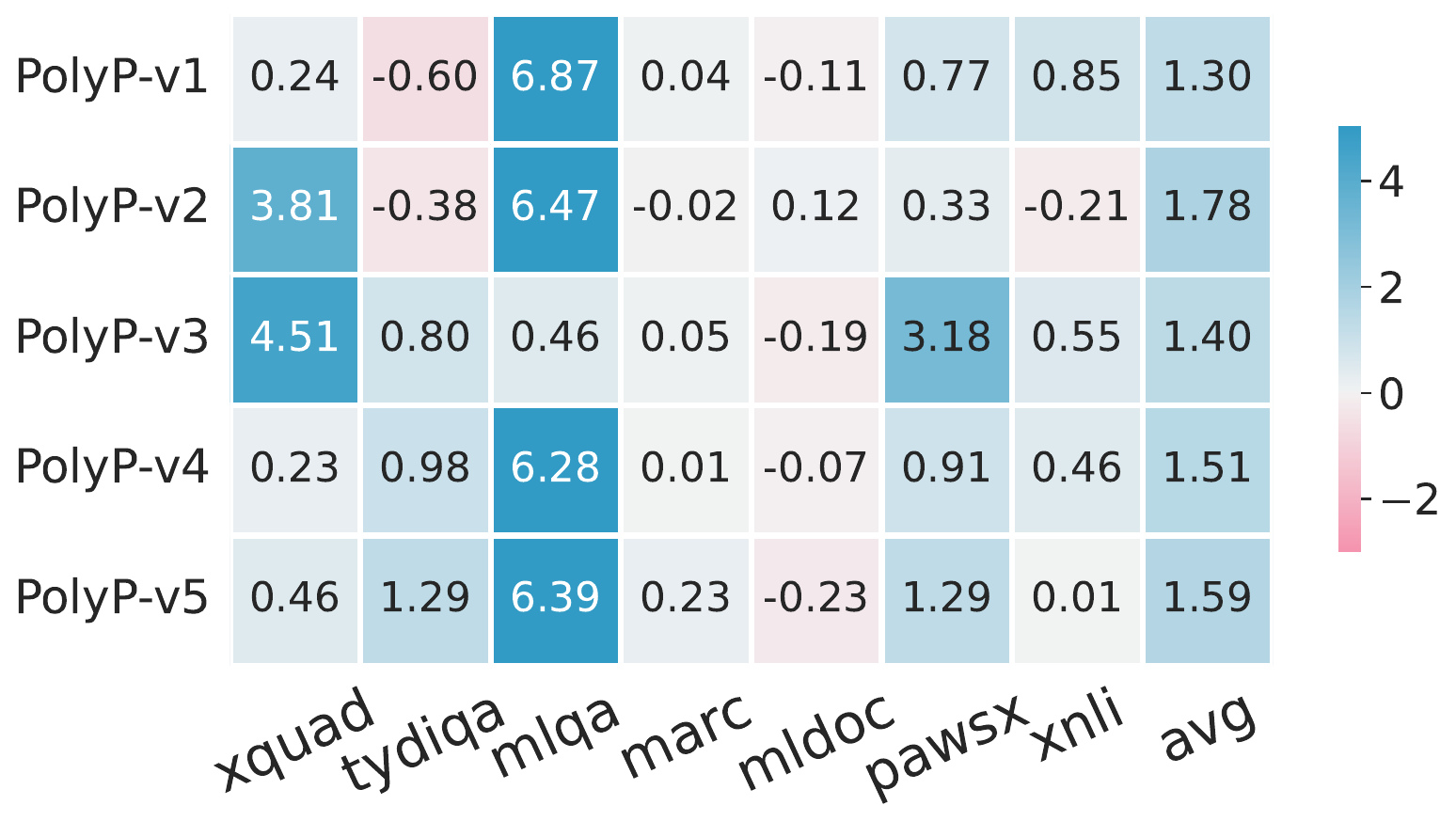}
	\end{minipage}
	}
	\vspace{-10pt}
	\caption{
	The exploration of the language and uniformity of prompt design. (a) is the performance gap between cross-lingual (CL) and in-lingual (IL) prompt templates, where \textit{PolyP}, \textit{PolyPE}, \textit{PolyPEX}, and \textit{PolyPEP} are abbreviations for \textit{PolyPrompt}, \textit{PolyPrompt+Expand}, \textit{PolyPrompt+Expand+XLSum}, and \textit{PolyPrompt+Expand+PANX}. 
	(b) is the relative performance improvement of \textit{PolyPrompt} with unified prompt templates versus diversified prompt templates (e.g. \textit{PolyP-v1}). \textit{PolyP-v(x)} ($x \in [1,5]$) represent $x$-th version of diversified prompt templates. 
	The bluer color indicates that the model with the cross-lingual (unified) prompts outperforms the in-lingual (diversified) prompts more, while the redder color has the opposite meaning. The last column is the average relative improvement.
}
	\label{fig:prompt-select}
\end{figure*}

\subsection{Exp-III: Effect of Prompt} 
\label{sec:exp3}

In this section, we try to find out what prompts or prompt combinations are suitable for multilingual and multitask scenarios (\textbf{Q3}).

\subsubsection{Prompt Design}
\label{sec:prompt}

Although prompting methods have proven effective in many NLP scenarios, its effectiveness comes at the cost of prompt engineering \cite{liu2021prompt}, as there are usually a multitude of factors that influence the prompt design process, and the situation is clearly more complicated in the multilingual situation.
Existing works have studied manual prompt \cite{timo2021exploiting}, soft (trainable) prompt \cite{brian2021power}, and mix prompt (mixing the manual and soft prompt) \cite{yuxian2021ppt,multilingual2021prompt} design approaches.
In this work, we take particular care of \textit{language} and \textit{uniformity} of prompt templates designed for multilingual multitask setting. 
The examples of the considered prompt design can be seen in Tab. \ref{tab:prompt-type}.

\noindent
\textbf{Language Choice: } we consider both the \textit{in-lingual} and \textit{cross-lingual} prompts. \textit{In-lingual prompts} are those in which the language of the prompt is the same as  the target language~\cite{multilingual2021prompt}. \textit{Cross-lingual prompts} denote those in which the language of the prompt template is different from the target language.  In this work,  we keep the language of the prompt template in English (en) \cite{victoria2021fewshot} regardless of the target language (e.g., zh). 

\noindent
\textbf{Uniformity of Templates:} 
Previous studies \cite{caruana1997multitask,evgeniouP04multitask,argyriou2008convex} have shown that similar tasks would benefit from multitask training. In this work, we study \textit{unified prompts} versus \textit{diversified prompts}. 
\textit{Unified prompts} indicates that prompt templates of different datasets have similar structures and cooccurrences.
\textit{Diversified prompts} means that the prompt templates for each task did not consider the same structure and multiple co-occurrence words. 
In practice, for each dataset, we designed $5$ different prompts and then randomly selected one prompt for each task to build a set of diverse prompts for multitask prompt training. In total, we created $5$ groups of diversified templates. The list of templates can be found in App.~\ref{sec:prompt-temp}.

 \subsubsection{Results}
\noindent
(1) \textbf{Cross-lingual prompt can help better retrieve knowledge encoded in language model.} 
We can observe from Fig.~\ref{fig:prompt-select}-(a) that the average overall performance of the $5$ models equipped with CL prompts outperformed IL prompts, which holds for all the seven datasets. We think they might be because mT5 was pre-trained on a larger body of English corpus,  it can understand the English template well. This makes it easier for downstream NLP tasks to retrieve knowledge from mT5.

\noindent 
(2) \textbf{The unified template outperforms the diversified template}
In Fig.~\ref{fig:prompt-select}-(b), we observed that the \textit{PolyPrompt} with uniform templates outperforms any diverse templates (e.g. \textit{PolyP-v1}), especially on the QA task.
The reason may be that unified prompts helped eliminate the boundaries between tasks, thereby reducing the distance between tasks and making the interaction between tasks easier.

\subsection{Exp-IV: Cross-task Cross-lingual zero-shot transfer}
\begin{figure*}[!t]
    \centering
    \includegraphics[width=0.75\linewidth]{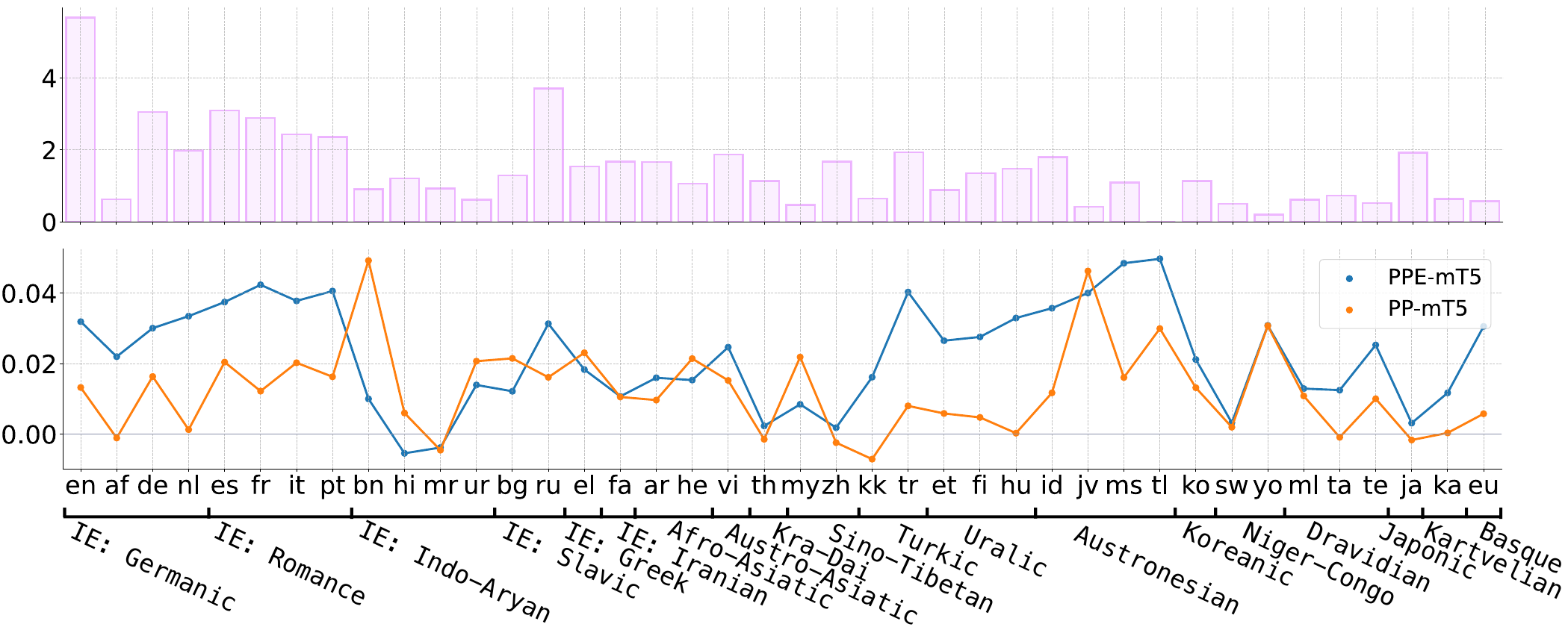}
    \vspace{-6pt}
    \caption{The performance improvement of \textit{PolyPrompt} (\textit{PP}, \textcolor{orange}{orange} line) and \textit{PolyPrompt+Expand} (\textit{PPE}, \textcolor{my_darkblue}{blue} line in line chart) relative to \textit{vanilla mT5} in the cross-task \& cross-lingual zero-shot setting.
    The bar chart is the proportion (\%) of different languages in the training corpus participating in mT5 pre-training, where \texttt{tl} is a language that never participated in mT5 pre-training. 
    }
    \label{fig:panx-zero}
\end{figure*}

To investigate whether \textit{PolyPrompt} is better at retrieving relevant knowledge from pre-trained language models for tasks and languages unseen in training stage, we investigate \textit{vanilla mT5}, \textit{PolyPrompt}, and \textit{PolyPrompt+Expand} fine-tuned on the English datasets and evaluate these three models on the PANX dataset, a named entity recognition task with 40 languages. 
We then subtract the performance of \textit{vanilla mT5} from \textit{PolyPrompt} and \textit{PolyPrompt+Expand} in the same language, and the results are shown in Fig.~\ref{fig:panx-zero}.

\paragraph{Results: }

\noindent
(1) \textbf{Almost all languages benefit from both \textit{PolyPrompt} and its variants.}
\textit{PolyPrompt}  brings gains for 34 of the 40 languages, and more languages will benefit when \textit{PolyPrompt} is enhanced with high-resource English training datasets.
Interestingly, \textit{PolyPrompt+Expand} performed much better than \textit{PolyPrompt} in languages belonging to \textit{IE: Germanic} and \textit{IE: Romance} language families, which made up a large proportion of samples in the pre-training corpus of mT5.

\noindent
(2)
\textbf{
\textit{PolyPrompt} significantly improves performance on languages that have never appeared in the pre-training corpus of \textit{mT5}.}
Both \textit{PolyPrompt} and \textit{PolyPrompt+Expand} improve a lot over mT5 on \texttt{tl}, a language that never appeared in mT5's pre-training corpus.
Furthermore, \textit{PolyPrompt+Expand} achieves the best performance gain on \texttt{tl}.
The reasons can be attributed to (1) we unify different tasks into a monolithic framework (including NER), which effectively shortens the distance between different tasks; (2) \texttt{English (en)} and \texttt{tl} share the same semantic space, NER knowledge in \texttt{English (en)} can be effectively transferred to \texttt{tl}.

\section{Conclusions}
\label{sec:conclusion}
We can provide the following preliminary empirical answers to our research questions.

\noindent
(1) \textit{\textbf{Can different tasks from different languages benefit from each other by a monolithic framework?}}
Yes. What's more, introducing more high-resource datasets can further improve the tasks' performance involved in multitask prompt training.

\noindent
(2) \textit{\textbf{How do different characteristics of datasets and languages affect the performance of \textit{PolyPrompt}?
}} 
\textit{PolyPrompt} cannot benefit all languages in all datasets. 
For example, (a) languages that appear only once in target datasets have benefits when \textit{PolyPrompt} is enhanced by high-resource datasets; (b) \textit{PolyPrompt} is better in short context samples for \text{MLQA}, long context samples for \text{XQuAD}, while poor in long question samples for \text{XQuAD}, \text{TyDiQA}, and \text{MLQA}.

\noindent
(3) \textit{\textbf{What makes a good prompt for multilingual multitask prompt training?}}
The best performance is achieved when the model is equipped with cross-lingual prompts (i.e., using English as prompt templates regardless of what the language of training samples is) and prompts with unified templates across tasks.

\section{Limitations}

Although in this paper, we try to cover as many languages and tasks as possible, some tasks (e.g., semantic parsing, machine translation) and languages are still not considered.
In addition, due to limited computational resources, we adopt a relatively small pre-trained language model, and the results on the larger pre-trained language models are also worth expecting. 
In addition, there are a variety of factors affecting the design of prompts in a multilingual setting. This paper only considers two (language choice and uniformity of prompt templates), so more comprehensive studies in this direction could be conducted.

\section*{Acknowledgements}
We thank Graham Neubig and Junjie Hu for their useful discussion and suggestions on this work.
This work was supported by the National Research Foundation of Singapore under its Industry Alignment Fund – Pre-positioning (IAF-PP) Funding Initiative. Any opinions, findings, conclusions, or recommendations expressed in this material are those of the authors and do not reflect the views of the National Research Foundation of Singapore.

\bibliographystyle{acl_natbib}
\bibliography{anthology,custom}

\appendix

\section{Languages}
\label{sec:language}
In this work, we studied $49$ languages that appear in $24$ datasets covering $6$ NLP tasks. For brevity, the languages are shown in ISO 639-1 codes\footnote{\url{https://en.wikipedia.org/wiki/List_of_ISO_639-1_codes}} as follows: af, am, ar, az, bg, bn, cy, de, el, en, es, et, eu, fa, fi, fr, gu, ha, he, hi, hu, id, ig, it, ja, jv, ka, kk, ko, ml, mr, ms, my, nl, np, pa, pt, ru, si, sw, ta, te, th, tl, tr, ur, vi, yo, zh. 
Among them, \textbf{zh, ja, th, te, km} are languages that do not use space separation for words.

\section{Datasets} 
\label{sec:dataset-app}

\subsection{Target Datasets} 
\label{sec:target-dataset-app}

\paragraph{XQuAD} \cite{DBLP:conf/acl/ArtetxeRY20}
is a cross-lingual question answering dataset, including 11 languages. 
Its English dataset is a subset of the development set of SQuAD v1.1. The other 10 languages are professional translations of the English dataset. Therefore, the dataset in 11 languages is completely parallel.

\paragraph{MLQA} \cite{DBLP:conf/acl/LewisORRS20}
is another multi-language extractive QA dataset, including 7 languages. Each QA instance is paralleled between 4 languages on average. Since MLQA and XQuAD lack training sets, following \citep{pmlr-v119-hu20b}, we use the training data of SQuAD v1.1 as their training set.

\paragraph{TyDiQA-GoldP (TyDiQA)}~\cite{DBLP:journals/tacl/ClarkPNCGCK20} is the gold passage version of the TyDiQA benchmark, including 9 languages for training, development, and testing. TyDiQA-GoldP is a simplified version of the primary task, discarding Thai and Japanese languages and samples without answers. Like XQuAD and MLQA,  
TyDiQA is evaluated with SQuAD 1.1 \cite{Pranav2016SQuAD2} metrics.  

\paragraph{XNLI}
\cite{DBLP:conf/emnlp/ConneauRLWBSS18} is a cross-lingual natural language inference dataset containing annotated development and test sets in 15 languages, and an annotated English training set. The English training set has been translated into the other 14 languages through a machine translator.\footnote{\url{https://console.cloud.google.com/storage/browser/xtreme_translations}}

\paragraph{PAWS-X} \cite{DBLP:conf/emnlp/YangZTB19}
 is a cross-lingual paraphrase adversary from a word scrambling dataset with 7 languages. The goal of this task is to predict whether two sentences are paraphrases. The training set of PAWS-X  is the PAWS's training data, and the subset of PAWS's development and test sets are translated into 6 other non-English datasets for evaluation.

\paragraph{MARC} (multilingual Amazon Reviews Corpus) \cite{keung2020multilingual} is a multilingual text classification dataset with 6 different languages. Here, we use the binarized classification task that is defined by \citet{keung2020multilingual}.

\paragraph{MLDOC} \cite{schwenk18mldoc} is a multilingual document classification dataset with six topics.

\subsection{Expanding Datasets}
\label{sec:expand-dataset-app}
Expanding tasks simply provide training sets for the multitask prompt training. 
In summary, we studied 15 English and 2 multilingual datasets.

\paragraph{Extractive Question Answering} is the task of finding an answer to a given question from the context. We adopt \text{SQuAD 2.0} \cite{Pranav2016SQuAD2}, \text{Quoref} \cite{Pradeep2019Quoref}, \text{NewsQA} \cite{Adam2017NewsQA}, and \text{ROPES} \cite{Kevin2019ropes}.

\paragraph{Multiple-choice Question Answering} aims to select an answer from candidate options based on the context and question.
In this work, we study \text{MCTest} \cite{Matthew2013MCTest} and \text{Social IQa} \cite{Maarten2019SocialIQA}.

\paragraph{Natural Language Inference} aims to determine the inference relation (e.g. entailment) between two texts. The datasets used in this work are \text{Quora},\footnote{\url{https://huggingface.co/datasets/quora}} \text{RTE} \cite{Alex2019GLUE}, and \text{SNLI} \cite{Samuel2015SNLI}.

\paragraph{Topic Classification} is a task to predict a suitable topic (e.g., health) for a given text. We use the following topic classification datasets: 
\text{DBpedia-2014} \cite{Xiang2015character},
\text{AG\_News} \cite{Xiang2015character}, and
\text{YATC} (Yahoo! Answers Topic Classification Dataset) \cite{Xiang2015character}.

\paragraph{Sentiment Classification} aims to identify the sentiment polarity of a given text. We studied datasets \text{IMDB} \cite{Andrew2011imdb}, \text{Amazon Review Polarity (ARP)} \cite{Xiang2015character}, and \text{SST2} \cite{Richard2013sst2}.

\paragraph{XL-Sum} \cite{tahmid2021xlsum} is a multilingual summarization dataset covering 45 low- to high-resource languages. 
We randomly select 32 out of 45 languages for multitask prompt training. The ISO-639-1 codes for the chosen languages are \textit{en, ar, vi, ko, es, zh, ru, fr, tr, hi, id, fa, pt, mr, th, az, bn, np, sr, sw, ta, te, ur, cy, am, my, gu, ha, ig, pa, si, yo}.

\paragraph{PANX} \cite{Xiaoman2017wikiann} is a multilingual named entity recognition dataset in 40 languages constructed based on Wikipedia corpus. Following \citet{pmlr-v119-hu20b}, we use the version with balanced train, development, and test splits from \citet{Afshin2019}.

\section{Interpretable Multilingual Evaluation}
\label{sec:interp-define-app}

For interpretable evaluation, the first step is attribute definition, and the second is sample breakdown.
Assume that $\phi_{\text{Len}}(x)$ is a function to calculate the number of tokens in the given text $x$, and $\phi_{\text{BLUE}}(x_1, x_2)$ is to compute the BLUE score of two given texts $x_1$ and $x_2$. The following are the features tailored for the $7$ multilingual datasets in this paper:

\begin{itemize*}
    \item \textbf{XQuAD, TyDiQA, MLQA}: \texttt{cLen}=$\phi_{\text{Len}}(X_{c})$, \texttt{qLen}=$\phi_{\text{Len}}(X_{q})$, \texttt{aLen}=$\phi_{\text{Len}}(X_{a})$, and \texttt{BLUE\_AC}= $\phi_{\text{BLUE}}(X_{a}, X_{c})$, where $X_{c}$, $X_q$, and $X_a$ denote the context, question, and answer sequence, respectively.
    \item \textbf{PAWS-X, XNLI}:  \texttt{t1Len} = $\phi_{\text{Len}}(X_{t1})$, \texttt{t2Len} = $\phi_{\text{Len}}(X_{t2})$,
    \texttt{t1Len/t2Len} = $\phi_{\text{Len}}(X_{t1}) / \phi_{\text{Len}}(X_{t2})$, and \texttt{BLUE\_t1t2} =  $\phi_{\text{BLUE}}(X_{t1}, X_{t2})$, where 
    $X_{t1}$ and $X_{t2}$ denote the premise and hypothesis  (sentence-1 and sentence-2 for PAWS-X) sequence.
    \item \textbf{MARC, MLDOC}: 
 \texttt{t1Len}=$\phi_{\text{Len}}(X_{t1})$, \texttt{t1basic} = $\phi_{\text{basic}}(X_{t1})$, and
    \texttt{t1eNum} = $\phi_{\text{eNum}}(X_{t1})$, where     $X_{t1}$ denotes a sequence of review (news for MLDOC). $\phi_{\text{basic}}(x)$ and $\phi_{\text{eNum}}(x)$ are functions to calculate the proportion of words belonging to the 1000 essential English words \footnote{\url{https://simple.wikipedia.org/wiki/Wikipedia:List_of_1000_basic_words}} and entities, respectively.
\end{itemize*}
We then follow \citet{fu2021interpteval} and breakdown the samples into four buckets, XS (extra-small), S (small), L (large), and XL (extra-large), according to their feature values, and calculate the performance for each bucket.

\begin{figure*}[ht]
	\centering
	\subfigure[CL-IL  \textit{PolyPrompt}]{
	\begin{minipage}[b]{0.31 \textwidth}
		\includegraphics[width=1\textwidth]{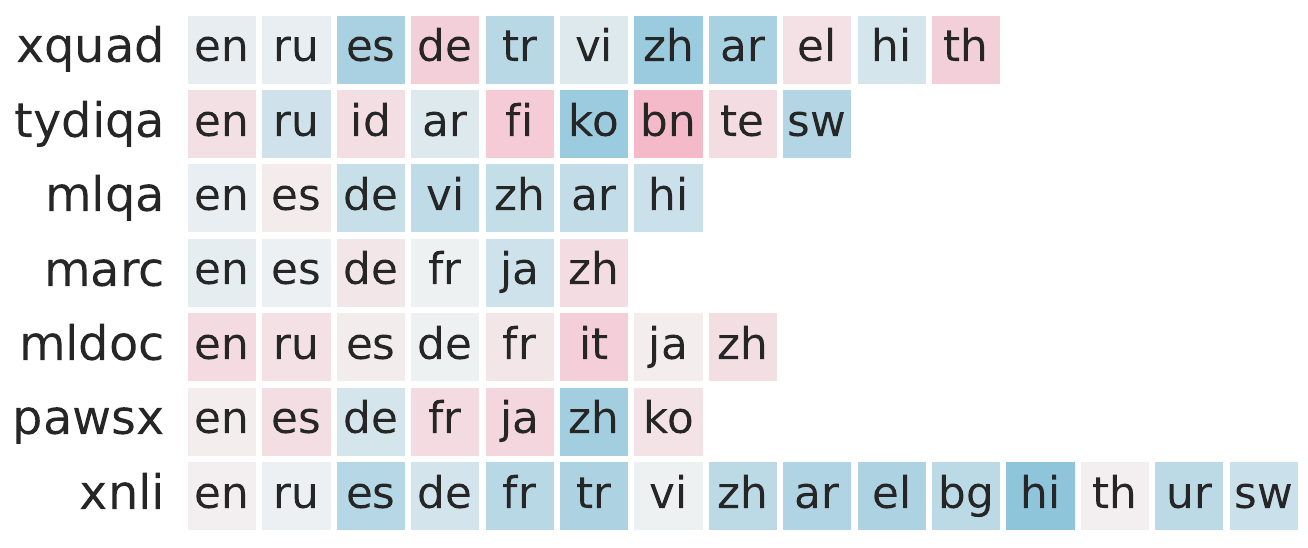}
	\end{minipage}
	} 
	\hspace{-8pt}
	\subfigure[CL-IL  \textit{PolyPrompt+Expand}]{
	\begin{minipage}[b]{0.27 \textwidth}
		\includegraphics[width=1\textwidth]{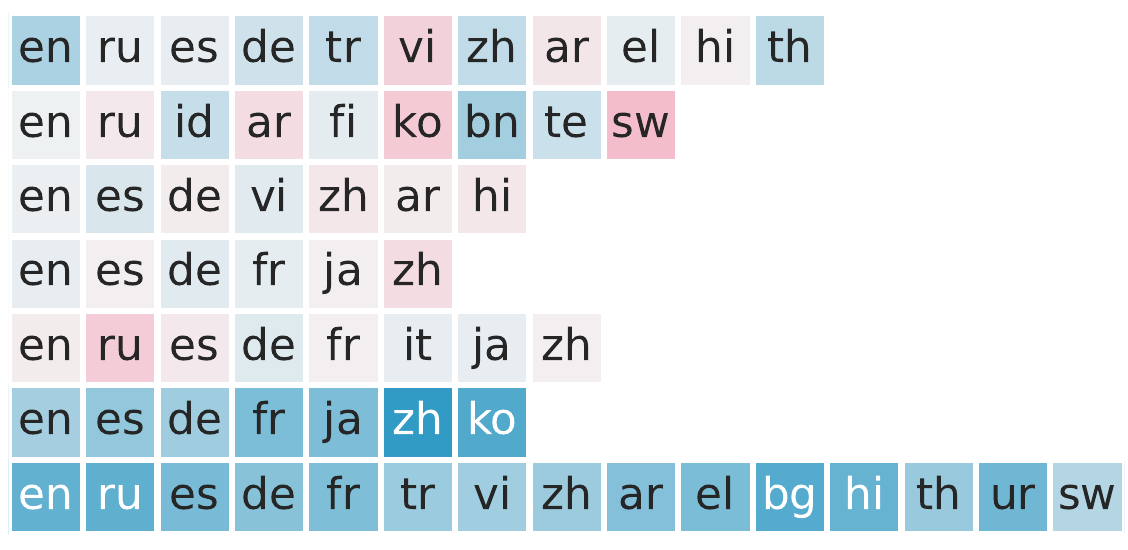}
	\end{minipage}
	}
	\hspace{-8pt}
	\subfigure[CL-IL  \textit{PolyPrompt+Expand+XLSum}]{
	\begin{minipage}[b]{0.31 \textwidth}
		\includegraphics[width=1\textwidth]{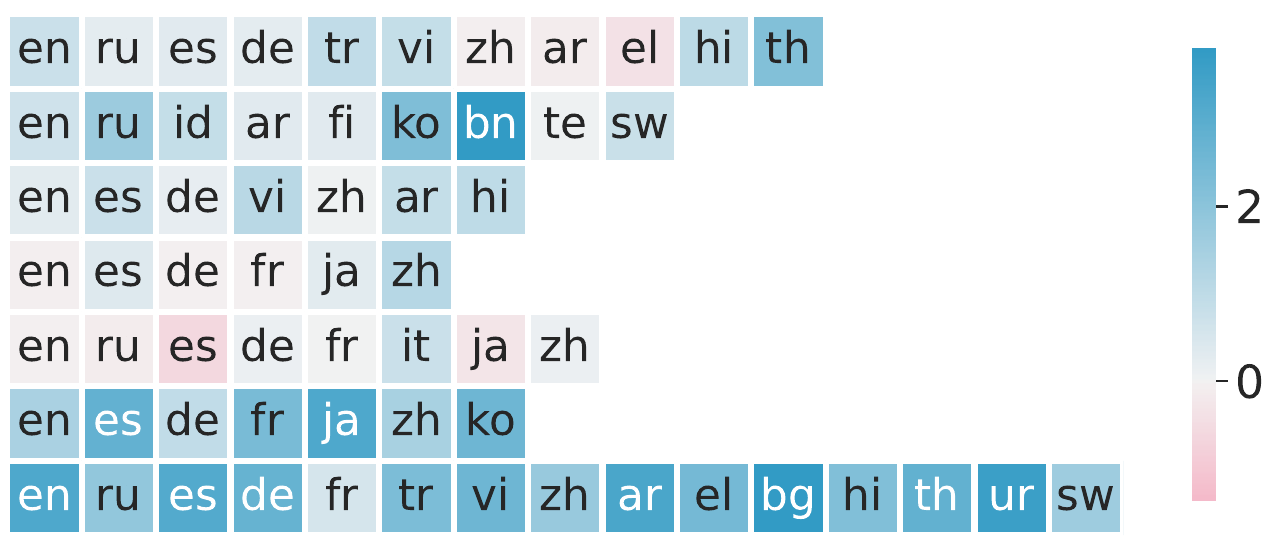}
	\end{minipage}
	} 
	\vspace{-7pt}
	\caption{
	The relative performance improvement at the language level for \textit{PolyPrompt} with cross-lingual prompts (CL) versus in-lingual prompts (IL).
Languages are sorted in descending order according to the sample size of the languages in the mT5 training set (high-resource to low-resource from left to right).
The bluer (redder)  the color, the greater the improvement (decrease) of CL over IL. 
}
	\label{fig:cl-il-app}
\end{figure*}

\section{Main Observations}
Due to the space limitation, we summarize some main observations here.

\noindent
\textbf{(1) Whether a language that appears in only one task could gain improvement depends on the difficulty of the task. }
In Fig.~\ref{fig:mtl-stl-family}-(a), we can observe that some languages in \text{XNLI}-[\texttt{ur,bg}], \text{TyDiQA}-[\texttt{bn,fi,id,te}] and \text{MLDOC}-[\texttt{it}] were not present in other tasks (e.g. \texttt{it} is only present in \text{MLDOC}).
These languages that appeared only once in multitask training have significant performance gains on the XNLI dataset, while performance dropsped significantly on the \text{TyDiQA} and \text{MLDOC} datasets.
The reason could be that XNLI is a task that relies more on fundamental knowledge \cite{yin-etal-2019-benchmarking}, which is relatively easier to acquire from other tasks. On the contrary, tasks such as \text{TyDiQA} need to understand more, for example, the semantics of sentences and the position of the answer.

\noindent
(2) \textbf{\textit{PolyPrompt} improves the performance of non-Indo-European languages a lot in the \textit{in-language training}.}
From Fig.~\ref{fig:mtl-stl-family}-(a), we can observe that  languages belonging to \textit{non-Indo-European} language families (e.g. \textit{Sino-Tibet} and \text{Niger-Congo}) always have performance gains no matter which datasets were employed.
However, in languages belonging to the \textit{Indo-European}-related language families, the relative performance gains varied widely across datasets.
For example, languages belonging to the XNLI and XQuAD datasets consistently achieved positive relative performance, while languages belonging to the PAWS-X and MLDOC datasets mainly achieved negative relative performance. However, this problem was found to be alleviated after introducing additional high-resource datasets (e.g. \textit{PolyPrompt}+Expand).

\begin{figure*}[th!]
  \centering \scriptsize
  \renewcommand\tabcolsep{0.3pt}
    \renewcommand\arraystretch{1.1}  
    \begin{tabular}{cccc cccc cccc cccc cccc cccc cccc cccc cccc cccc}
    \toprule
 \multicolumn{8}{c}{\textbf{PAWS-X}} & 
 \multicolumn{8}{c}{\textbf{XNLI}} & 
 \multicolumn{8}{c}{\textbf{XQuAD}} & 
 \multicolumn{8}{c}{\textbf{TyDiQA}} & 
 \multicolumn{8}{c}{\textbf{MLQA}} \\
\cmidrule(lr){1-8}\cmidrule(lr){9-16}\cmidrule(lr){17-24}\cmidrule(lr){25-32}\cmidrule(lr){33-40}
    \multicolumn{4}{c}{\texttt{t1Len}} & \multicolumn{4}{c}{\texttt{t2Len}} &
    \multicolumn{4}{c}{\texttt{t1Len}} & \multicolumn{4}{c}{\texttt{t2Len}} &
    \multicolumn{4}{c}{\texttt{cLen}} & \multicolumn{4}{c}{\texttt{qLen}} &
    \multicolumn{4}{c}{\texttt{cLen}} & \multicolumn{4}{c}{\texttt{qLen}} &
    \multicolumn{4}{c}{\texttt{cLen}} & \multicolumn{4}{c}{\texttt{qLen}} \\
\cmidrule(lr){1-8}\cmidrule(lr){9-16}\cmidrule(lr){17-24}\cmidrule(lr){25-32}\cmidrule(lr){33-40}
    \multicolumn{4}{c}{\multirow{5}[2]{*}{\includegraphics[scale=0.13]{fig/databias/pawsx_t1Len.pdf}}} & 
    \multicolumn{4}{c}{\multirow{5}[2]{*}{\includegraphics[scale=0.13]{fig/databias/pawsx_t2Len.pdf}}} & 
    \multicolumn{4}{c}{\multirow{5}[2]{*}{\includegraphics[scale=0.13]{fig/databias/xnli_t1Len.pdf}}} &
    \multicolumn{4}{c}{\multirow{5}[2]{*}{\includegraphics[scale=0.13]{fig/databias/xnli_t2Len.pdf}}} &
    \multicolumn{4}{c}{\multirow{5}[2]{*}{\includegraphics[scale=0.13]{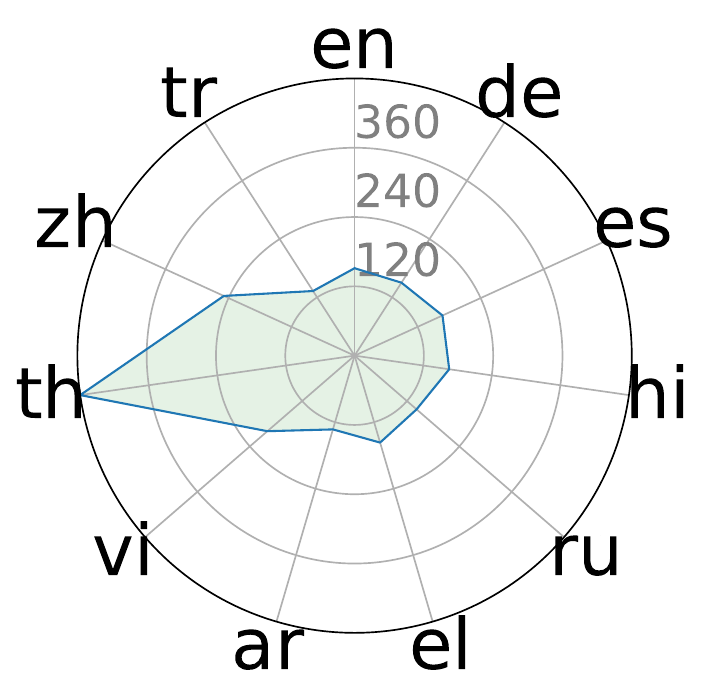}}} & 
    \multicolumn{4}{c}{\multirow{5}[2]{*}{\includegraphics[scale=0.13]{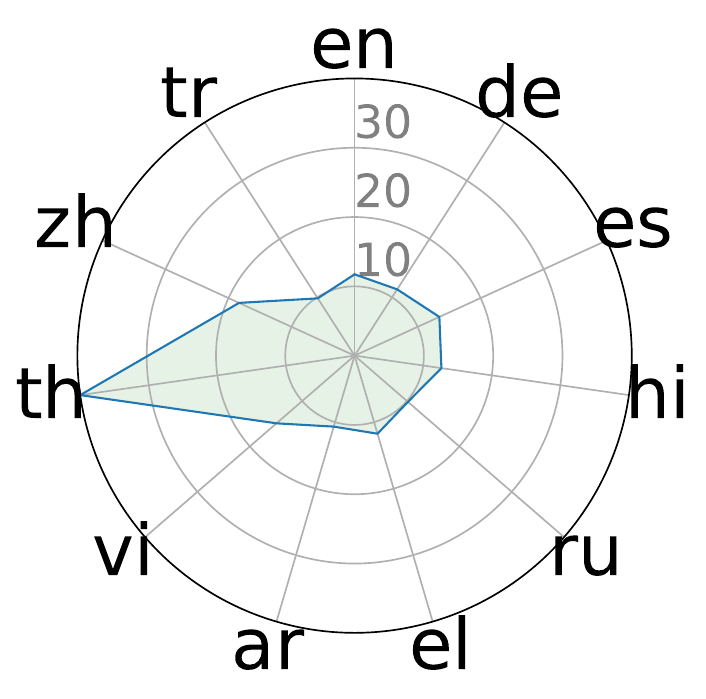}}} &
    \multicolumn{4}{c}{\multirow{5}[2]{*}{\includegraphics[scale=0.13]{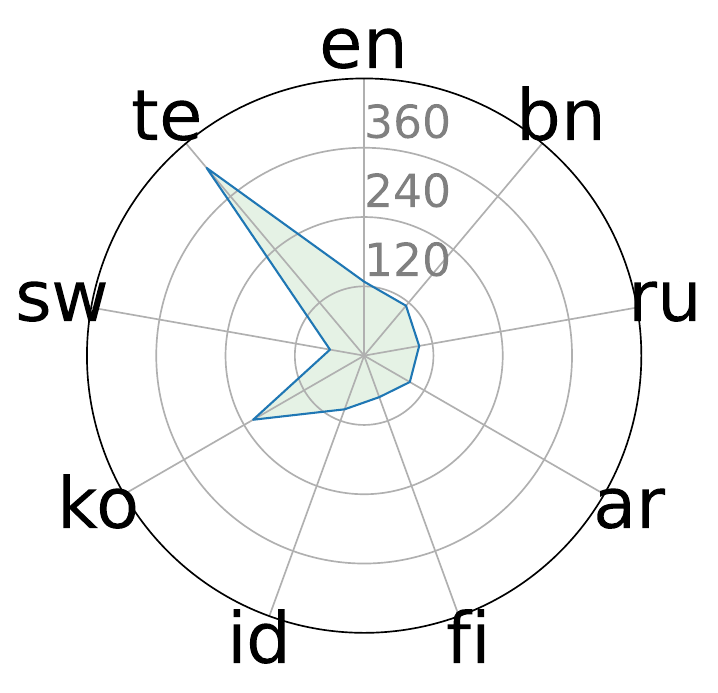}}} & 
    \multicolumn{4}{c}{\multirow{5}[2]{*}{\includegraphics[scale=0.13]{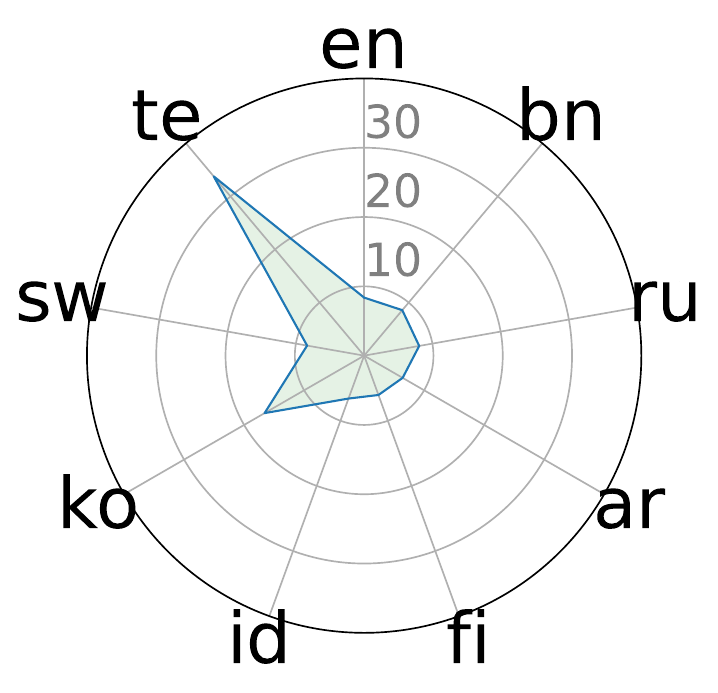}}} &
    \multicolumn{4}{c}{\multirow{5}[2]{*}{\includegraphics[scale=0.13]{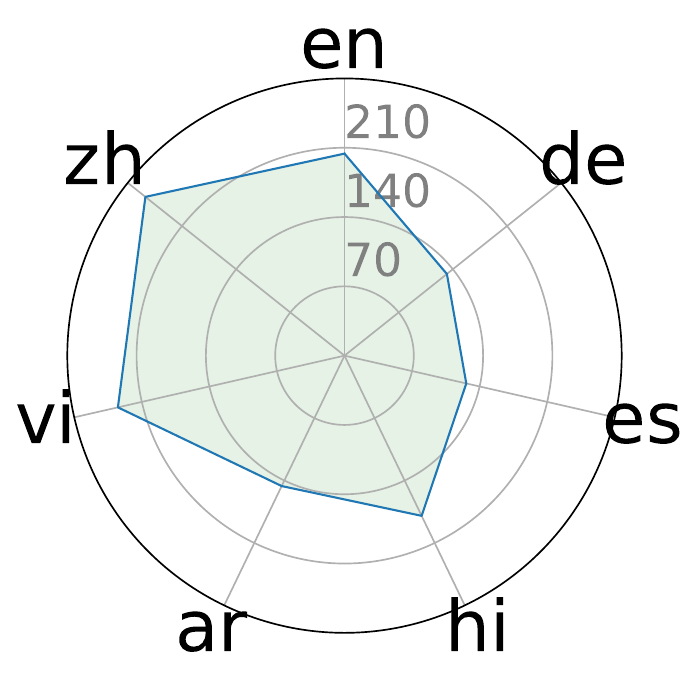}}}  &
    \multicolumn{4}{c}{\multirow{5}[2]{*}{\includegraphics[scale=0.13]{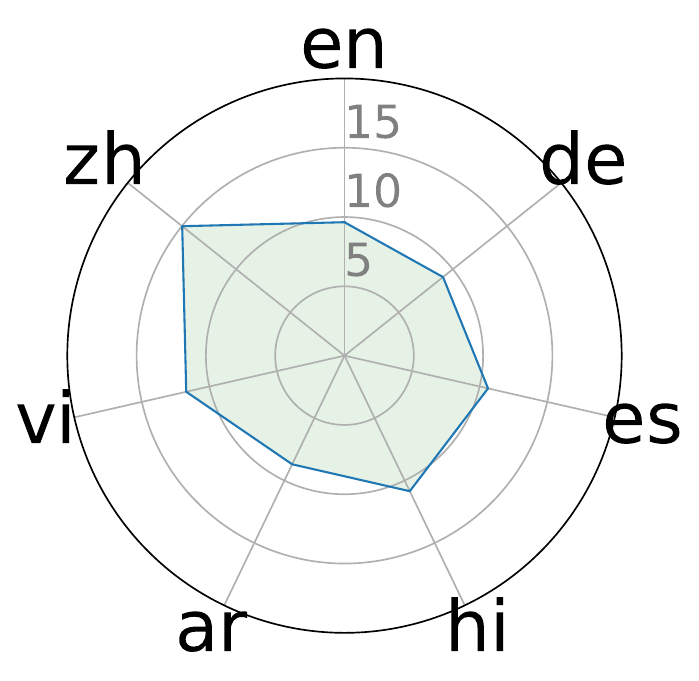}}}
    \\ \\ \\ \\ \\ 

 \cmidrule(lr){1-8}\cmidrule(lr){9-16}\cmidrule(lr){17-24}\cmidrule(lr){25-32}\cmidrule(lr){33-40}
     \multicolumn{4}{c}{\texttt{t1Len/t2Len}} & \multicolumn{4}{c}{\texttt{BLUE\_AC}} & 
     \multicolumn{4}{c}{\texttt{t1Len/t2Len}} & \multicolumn{4}{c}{\texttt{BLUE\_AC}} & 
     \multicolumn{4}{c}{\texttt{aLen}} & \multicolumn{4}{c}{\texttt{BLUE\_AC}} & 
     \multicolumn{4}{c}{\texttt{aLen}} & \multicolumn{4}{c}{\texttt{BLUE\_AC}} & 
     \multicolumn{4}{c}{\texttt{aLen}} & \multicolumn{4}{c}{\texttt{BLUE\_AC}}  \\
\cmidrule(lr){1-8}\cmidrule(lr){9-16}\cmidrule(lr){17-24}\cmidrule(lr){25-32}\cmidrule(lr){33-40}
    \multicolumn{4}{c}{\multirow{5}[2]{*}{\includegraphics[scale=0.13]{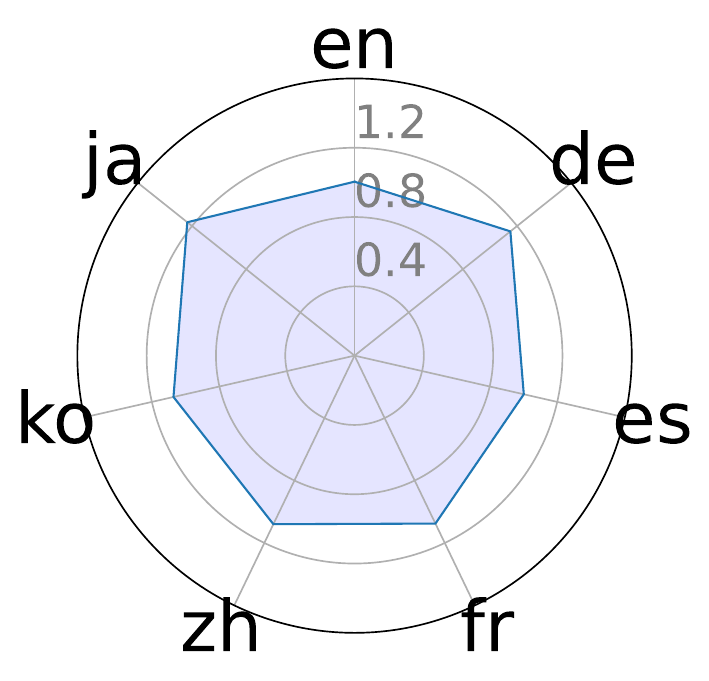}}} & 
    \multicolumn{4}{c}{\multirow{5}[2]{*}{\includegraphics[scale=0.13]{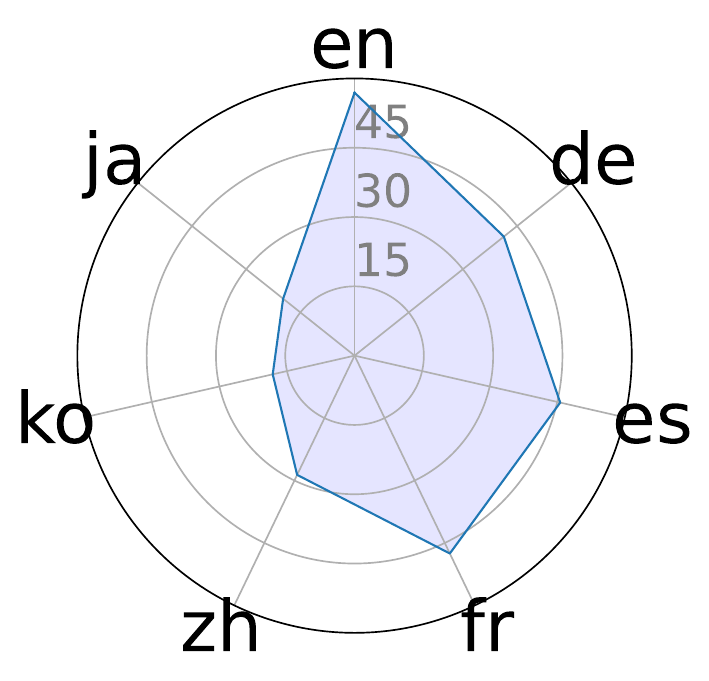}}} &
    \multicolumn{4}{c}{\multirow{5}[2]{*}{\includegraphics[scale=0.13]{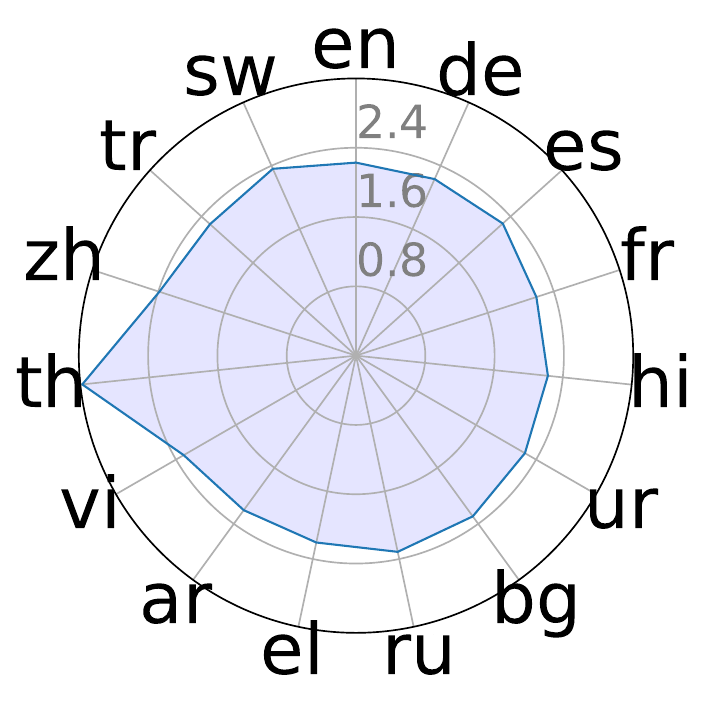}}} &
    \multicolumn{4}{c}{\multirow{5}[2]{*}{\includegraphics[scale=0.13]{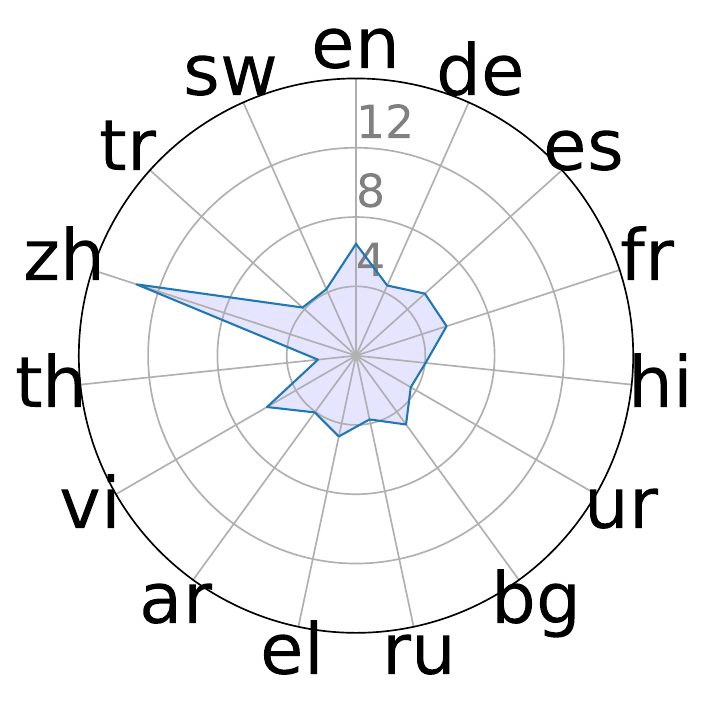}}} &
    \multicolumn{4}{c}{\multirow{5}[2]{*}{\includegraphics[scale=0.13]{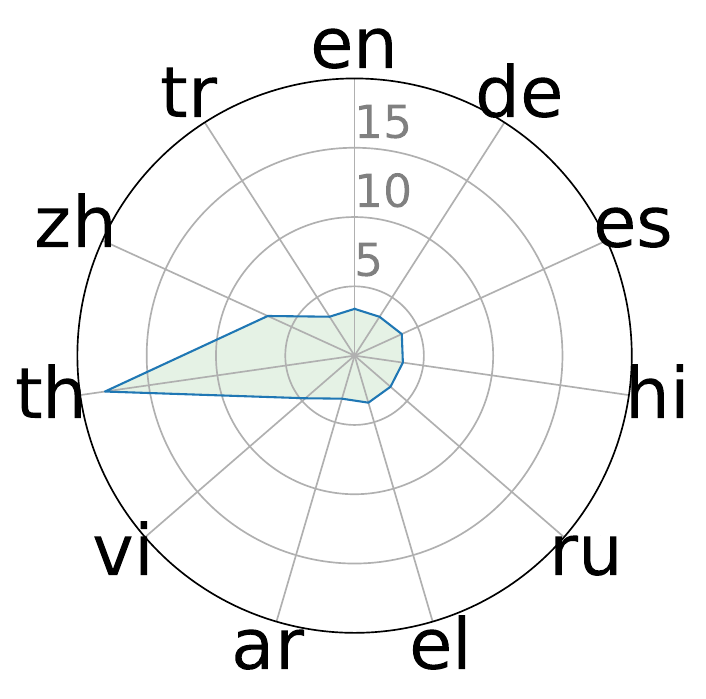}}} &
    \multicolumn{4}{c}{\multirow{5}[2]{*}{\includegraphics[scale=0.13]{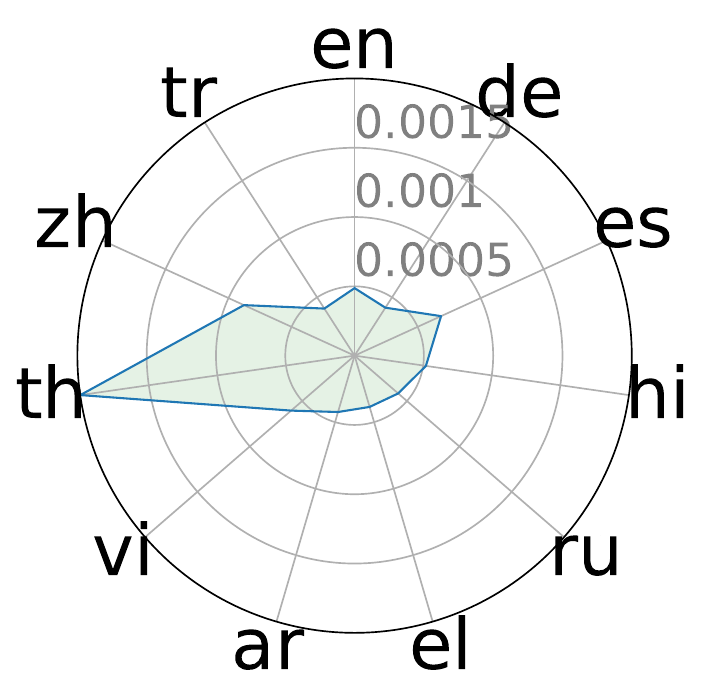}}} & 
    \multicolumn{4}{c}{\multirow{5}[2]{*}{\includegraphics[scale=0.13]{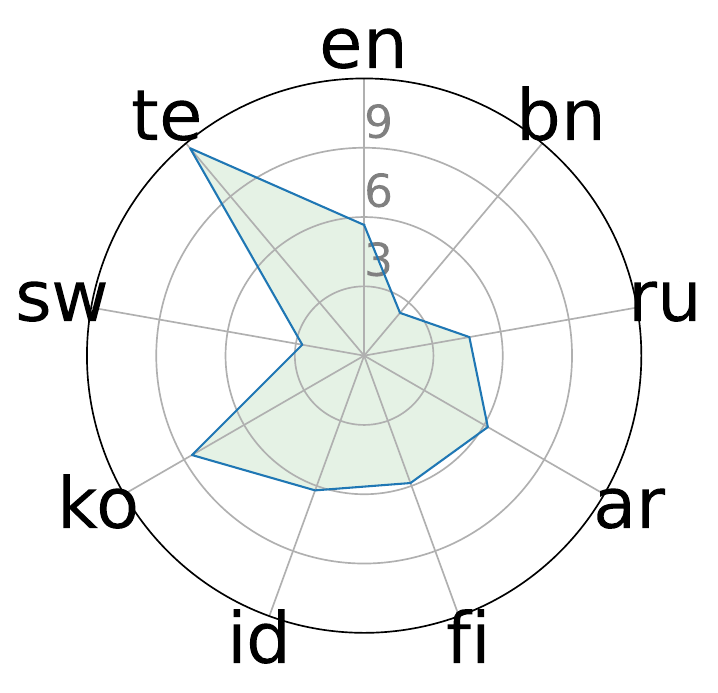}}} &
    \multicolumn{4}{c}{\multirow{5}[2]{*}{\includegraphics[scale=0.13]{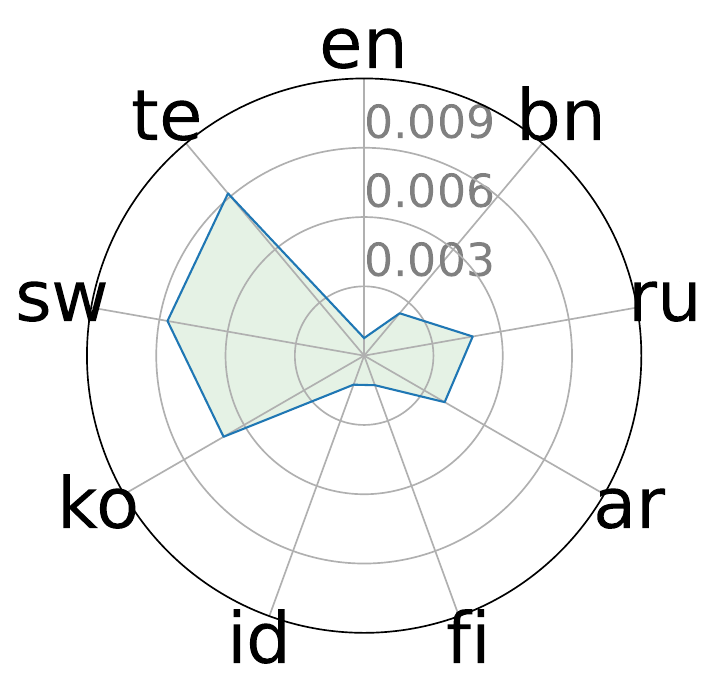}}} &
    \multicolumn{4}{c}{\multirow{5}[2]{*}{\includegraphics[scale=0.13]{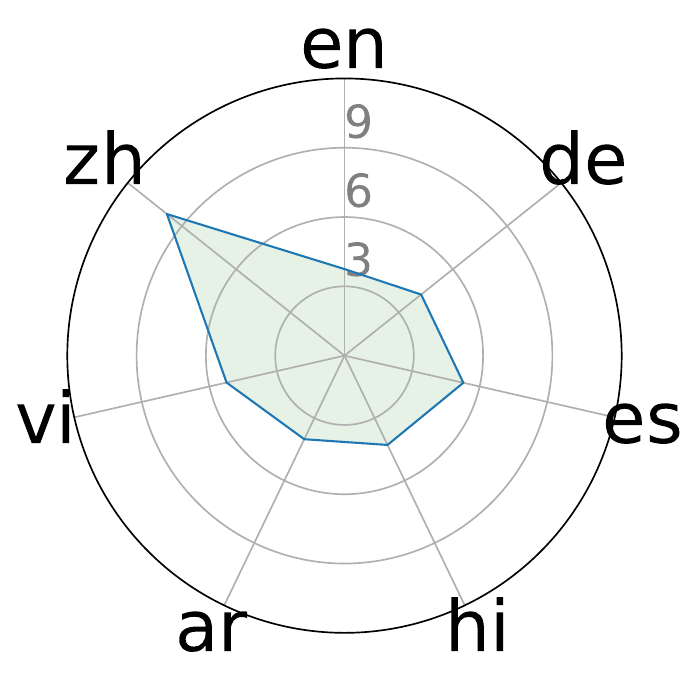}}} &
    \multicolumn{4}{c}{\multirow{5}[2]{*}{\includegraphics[scale=0.13]{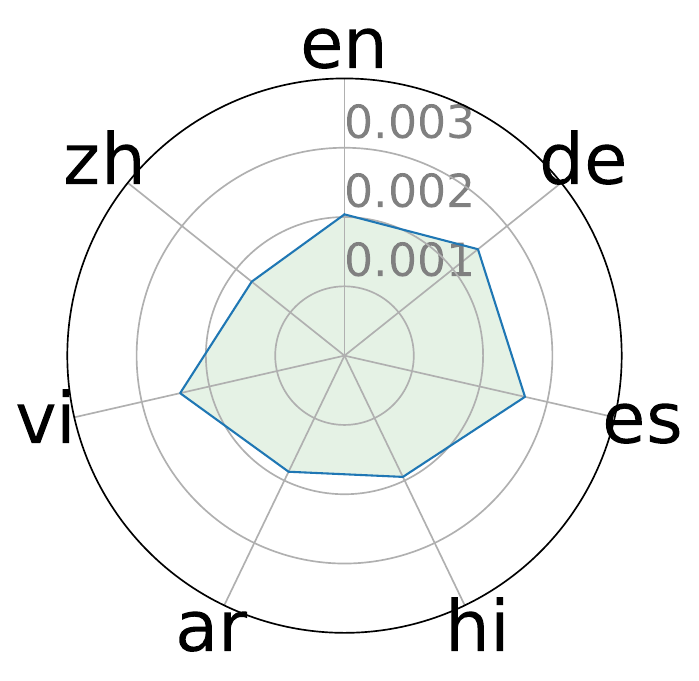}}} 
    \\ \\ \\ \\ \\
    \bottomrule 
    \end{tabular}
      \caption{Dataset bias characterized by $\phi_{p}$ defined in Eq.~\ref{eq:data-feat-app} 
      (the average of feature values over a specific language of the dataset).
      }
  \label{fig:data-bias-app}%
\end{figure*}%

\noindent
\textbf{(3) For low-resource languages, \textit{PolyPrompt} with in-language prompts will bring more gains, while cross-lingual prompts bring more gains when introducing high-resource training datasets.}
From Fig.~\ref{fig:cl-il-app}, we can observe that \textit{PolyPrompt} with in-lingual prompts outperform with cross-language prompts in low-resource languages. However, when the external English dataset was introduced (\textit{PolyPrompt+Expand}), cross-language prompts have more gains in both low and high resource languages. With the introduction of multilingual datasets (\textit{PolyPrompt+Expand+XLSum}), the relative advantages of cross-lingual prompts increased.

\section{Dataset Bias}
\label{sec:data-feat-define-app}
\paragraph{Dataset-level Features}
We also obtain the dataset-level features. Given a dataset $D$ and a feature $p$ (e.g. \texttt{qLen}), the dataset-level feature can be defined as: 
\begin{equation}
\label{eq:data-feat-app}
    \phi_{p}(D) = \frac{1}{|D^{te}|} \sum_{d\in D^{te}} \phi_{p}(d),
\end{equation}
where $d$ is a sample of the test set $D^{te} \in D$, and $\phi_{p}(\cdot)$ is a function that computes the feature value for a given sample. 
For example, $\phi_{\text{qLen}}(\text{MLQA})$ denotes the average question length of the \texttt{MLQA}.

Dataset bias is measured by $\phi_{p}$, the dataset-level feature defined in Eq.~\ref{eq:data-feat-app}. 
Tab.~\ref{fig:data-bias-app} shows five target datasets explored in Sec.~\ref{sec:interpret}.

\section{Prompt Template}
\label{sec:prompt-temp}
Tab.~\ref{tab:all-prompt} presents the cross-lingual (English) prompt templates explored in this work. We designed $5$ templates for each of the $7$ tasks.

\begin{table*}[htb]
  \centering \footnotesize
    \begin{tabular}{cm{13cm}}
    \toprule
    \textbf{Task} & \multicolumn{1}{c}{\textbf{Prompt Template}} \\
    \midrule
    \multirow{5}[2]{*}{\textbf{XQuAD}} & (1) Answer the question based on the paragraph. | Question: \texttt{[Q]} | Paragraph: \texttt{[C]} \\
          & (2) Answer the question based on the information contained in the paragraph. | Question: \texttt{[Q]} | Paragraph: \texttt{[C]} \\
          & (3) \texttt{[C]} | With reference to the above context, \texttt{[Q]} \\
          & (4) I have always wondered: \texttt{[Q]} | I searched Wikipedia and this is what I found. What's the answer? | \texttt{[C]} \\
          & (5) Context: \texttt{[C]} | I am trying to figure out the answer to the question from the above context. Can you tell me the answer? | Question: \texttt{[Q]} Answer: \\
    \midrule
    \multirow{5}[2]{*}{\textbf{TyDiQA}} & (1) Answer the question based on the paragraph. | Question: \texttt{[Q]} | Paragraph: \texttt{[C]} \\
          & (2) Answer the question based on the information contained in the paragraph. | Question: \texttt{[Q]} | Paragraph: \texttt{[C]} \\
          & (3) I have always wondered: \texttt{[Q]} | I searched Wikipedia and this is what I found. What's the answer? | \texttt{[C]} \\
          & (4) \texttt{[C]} | With reference to the above context, \texttt{[Q]} \\
          & (5) Context: \texttt{[C]} | I am trying to figure out the answer to the question from the above context. Can you tell me the answer? | Question: \texttt{[Q]} Answer: \\
    \midrule
    \multirow{5}[2]{*}{\textbf{MLQA}} & (1) Answer the question based on the paragraph. | Question: \texttt{[Q]} | Paragraph: \texttt{[C]} \\
          & (2) Answer the question based on the information contained in the paragraph. | Question: \texttt{[Q]} | Paragraph: \texttt{[C]} \\
          & (3) Context: \texttt{[C]} | I am trying to figure out the answer to the question from the above context. Can you tell me the answer? | Question: \texttt{[Q]} Answer: \\
          & (4) \texttt{[C]} | With reference to the above context, \texttt{[Q]} \\
          & (5) I have always wondered: \texttt{[Q]} | I searched Wikipedia and this is what I found. What's the answer? | \texttt{[C]} \\
    \midrule
    \multirow{5}[2]{*}{\textbf{XNLI}} & (1) Answer the question based on paragraph 1 and paragraph 2. | Question: Do Paragraph 1 and Paragraph 2 mean the same thing? | (A) Yes. (B) No. (C) Maybe. | Paragraph 1: \texttt{[T1]} | Paragraph 2: \texttt{[T2]} \\
          & (2) \texttt{[T1]} | Based on the previous passage, is it true that \texttt{[T2]}? Yes, no, or maybe? \\
          & (3) Suppose \texttt{[T1]} Can we infer that \texttt{[T2]}? | Option: (A) Yes (B) No (C) Maybe? \\
          & (4) Paragraph: \texttt{[T1]} | Question: Does this imply that ``\texttt{[T2]}''? ｜ Yes, no, or maybe? \\
          & (5) Given that \texttt{[T1]} Therefore, it must be true that \texttt{[T2]}? Yes, no, or maybe? \\
    \midrule
    \multirow{5}[2]{*}{\textbf{PAWS-X}} & (1) Answer the question based on paragraph 1 and paragraph 2. | Question: Do Paragraph 1 and Paragraph 2 express the same meaning? ｜ (A) Yes. (B) No. ｜ Paragraph 1: \texttt{[T1]} | Paragraph 2: \texttt{[T2]} \\
          & (2) Paragraph 1: \texttt{[T1]} | Paragraph 2: \texttt{[T2]} | Question: Do Paragraph 1 and Paragraph 2 express the same meaning? ｜ Yes or No? \\
          & (3) Suppose \texttt{[T1]} Can we infer that \texttt{[T2]}? | Option: (A) Yes (B) No? \\
          & (4) \texttt{[T1]} | Based on the previous passage, is it true that \texttt{[T2]}? Yes or no? \\
          & (5) Given that \texttt{[T1]} Therefore, it must be true that \texttt{[T2]}? Yes or no? \\
    \midrule
    \multirow{5}[2]{*}{\textbf{MARC}} & (1) Answer the question based on the review. | Question: Can we conclude that the buyer is satisfied with the product based on his review? ｜ (A) Yes. (B) No. | Review: \texttt{[T1]} \\
          & (2) I am reading a review that says \texttt{[T1]}. Do you think the review is positive or negative? \\
          & (3) Review: \texttt{[T1]} | Did the reviewer find this product good or bad? \\
          & (4) Review: \texttt{[T1]} | Is this review positive or negative? \\
          & (5) Review: \texttt{[T1]} | Is the buyer satisfied with the product purchased? \\
    \midrule
    \multirow{5}[2]{*}{\textbf{MLDOC}} & (1) Answer the question based on the news. | Question: What topic does this news belong to? | (A) Business and Industry. (B) Economy. (C) Government and Society. (D) Market. | News: \texttt{[T1]} \\
          & (2) Is this a piece of news regarding (A) Business and Industry (B) Economy (C) Government and Society (D) Market? | \texttt{[T1]} \\
          & (3) Article: \texttt{[T1]} | Which of the following sections of a newspaper would this article likely appear in? | Options: (A) Business and Industry (B) Economy (C) Government and Society (D) Market \\
          & (4) What topic does this news belong to? | (A) Business and Industry. (B) Economy. (C) Government and Society. (D) Market. | News: \texttt{[T1]} \\
          & (5) Would you recommend the following article to a (A) Business and Industry (B) Economy (C) Government and Society (D) Market? | \texttt{[T1]} \\
    \bottomrule
    \end{tabular}
    \caption{The cross-lingual (English) prompt templates studied in this work. ``\texttt{Q}'', ``\texttt{C}'', ``\texttt{T1}'', and ``\texttt{T2}'' denotes the placeholders for question, context, $\text{sentence}_1$, and $\text{sentence}_2$ field, respectively.}
  \label{tab:all-prompt}
\end{table*}

\end{CJK*} 
\end{document}